\RequirePackage{fix-cm}

\documentclass[smallcondensed]{svjour3}
\smartqed

\usepackage{hyperref}       
\usepackage{url}            
\usepackage{booktabs}       
\usepackage{amsfonts}       
\usepackage{nicefrac}       
\usepackage{microtype}      
\usepackage{amssymb}
\usepackage{amsmath}
\usepackage{mathtools}
\usepackage{algorithm}
\usepackage{algpseudocode}
\usepackage{ifthen}
\usepackage{subcaption}
\usepackage{xcolor}
\usepackage{bbm}
\usepackage{bm}
\usepackage{hyperref}

\DeclarePairedDelimiter{\ceil}{\lceil}{\rceil}

\newcommand{\calk}{\mathcal{K}}
\newcommand{\C}{\mathcal{C}}

\newcommand{\gest}[1]{\hat{g}_{#1,n}}

\newcommand{\prob}[1]{{\mathsf{Pr}}\left( #1 \right)}
\newcommand{\probnu}[2]{\mathbb{P}_{#1}\left( #2 \right)}

\newcommand{\remove}[1]{}

\newcommand{\Exp}[1]{\mathbb{E}\left[#1\right]}
\newcommand{\Expnu}[2]{\mathbb{E}_{#1}\left[#2\right]}

\DeclareMathOperator*{\argmin}{arg\,min}
 
\newcommand{\R}{\mathbb{R}}

\newcommand{\ignore}[1]{}

\newcommand{\lcb}{\text{LCB}}
\newcommand{\KL}{\text{KL}}
\newcommand{\Rinf}{R^{\mathsf{inf}}}
\newcommand{\Rsub}{R^{\mathsf{sub}}}

\newcommand{\Rcon}{R^{\mathsf{con}}}

\newcommand{\Xbf}{\textbf{X}}

\newboolean{showcomments}
\setboolean{showcomments}{true}
\newcommand{\ak}[1]{  \ifthenelse{\boolean{showcomments}}
{ \textcolor{red}{(AK says:  #1)}} {}  }
\newcommand{\jk}[1]{  \ifthenelse{\boolean{showcomments}}
{ \textcolor{red}{(JK says:  #1)}} {}  }
\newcommand{\kj}[1]{  \ifthenelse{\boolean{showcomments}}
{ \textcolor{red}{(KJ says:  #1)}} {}  }
\newcommand{\addcites}[0]{\ifthenelse{\boolean{showcomments}}
{ \textcolor{green}{(add citation(s))}}{}}
\newcommand{\addref}[0]{\ifthenelse{\boolean{showcomments}}
{ \textcolor{green}{(add ref)}}{}}
\newcommand{\revision}[1]{{#1}}

\usepackage[round]{natbib}
\begin{document}





\title{Constrained regret minimization for multi-criterion multi-armed
bandits}

\author{Anmol  Kagrecha         \and
        Jayakrishnan Nair \and
        Krishna Jagannathan
}


\institute{A. Kagrecha \at
              Stanford University \\
              \email{akagrecha@gmail.com}           
           \and
           J. Nair \at
              IIT Bombay \\
              \email{jayakrishnan.nair@ee.iitb.ac.in}
           \and
           K. Jagannathan \at IIT Madras \\
           \email{krishnaj@ee.iitm.ac.in}
}

\maketitle

\begin{abstract}
We consider a stochastic multi-armed bandit setting and study the
problem of constrained regret minimization over a given time
horizon. Each arm is associated with an unknown, possibly
multi-dimensional distribution, and the merit of an arm is determined
by several, possibly conflicting attributes. The aim is to optimize a
`primary’ attribute subject to user-provided constraints on other
`secondary’ attributes. We assume that the attributes can be estimated
using samples from the arms’ distributions, and that the estimators
enjoy suitable concentration properties. We propose an algorithm
called \textsc{Con-LCB} that guarantees a logarithmic regret, i.e.,
the average number of plays of all non-optimal arms is at most
logarithmic in the horizon. The algorithm also outputs a boolean flag
that correctly identifies, with high probability, whether the given
instance is feasible/infeasible with respect to the constraints. We
also show that \textsc{Con-LCB} is optimal within a universal
constant, i.e., that more sophisticated algorithms cannot do much
better universally. Finally, we establish a fundamental trade-off
between regret minimization and feasibility identification. Our
framework finds natural applications, for instance, in financial
portfolio optimization, where \emph{risk constrained} maximization of
expected return is meaningful.
\keywords{Multi-criterion multi-armed bandits \and constrained bandits 
\and regret minimization}
\end{abstract}

\section{Introduction}

The multi-armed bandit (MAB) problem is a fundamental construct in
online learning, where a learner has to quickly identify the best
option (a.k.a., arm) among a given set of options. In the stochastic
MAB problem, each arm is associated with an (a priori unknown) reward
distribution, and a sample from this distribution is revealed each
time an arm is chosen (a.k.a., pulled). The classical goal is to use
these samples to quickly identify the arm with the highest mean
reward. The most popular metric to evaluate the performance of a
learning algorithm is \emph{regret}, which captures how often a
suboptimal arm was pulled by the algorithm.

While the classical stochastic MAB formulation has been applied in
various application scenarios, including clinical trials, portfolio
optimization, anomaly detection, and telecommunication
\cite{Bouneffouf19}, it ignores a key aspect of most real-world
decision-making problems---namely, that \emph{they have multiple criteria of
  interest}. For example, when comparing testing kits in a clinical 
  trial, one would want to keep track of the false-positive rate as well
as the false-negative rate of each kit. Similarly, choosing the best
financial portfolio involves balancing risk and reward. A wireless
node deciding which channel to transmit on, or what transmission rate
to use has to balance several criteria, including throughput, delay,
and energy consumption. This multi-criterion nature of decision making
is not always adequately captured by the classical MAB approach of
optimizing a one-dimensional reward signal.

The most common approach for incorporating multiple arm attributes
into MAB formulations is to define the reward as a suitable function
(say a linear combination) of the attributes of interest. For example,
risk-aware portfolio optimization can be cast as an MAB problem where
the best arm is one that optimizes a certain linear combination of
mean value and a suitable risk measure (such as standard deviation or Conditional
Value at Risk) (see, for example, \cite{sani2012,
  vakili2016,kagrecha2019}. However, this approach assumes that the
different attributes of interest can be expressed and compared on a
common scale, which is not always reasonable. For example, how does
one `equate' the impact a certain increment in risk to that of an
another increment in the mean return of a portfolio?

A more natural approach for multi-criterion decision making is to
instead pose the optimal choice as the solution of a \emph{constrained
  optimization}. In other words, optimize one attribute subject to
constraints on the others. However, despite the modest literature
on multi-criterion MABs (surveyed below), little attention has been
paid to formulating, and designing algorithms for \emph{constrained}
multi-criterion MABs. This paper seeks to fill this gap. Formally, we
assume that each arm is associated with a $D$-dimensional probability
distribution, and the best arm is the one that minimizes a certain arm
attribute subject to constraints on~$m$ other attributes. For this
setting, we pursue regret minimization, i.e., we seek to minimize the
average number of pulls of non-optimal arms.

To simplify the presentation, and to highlight the key aspects of this
problem, we first consider a single constraint (i.e., $m = 1$). Each
arm is associated with a probability distribution, and we consider the
problem of optimizing an \emph{objective attribute}, subject to a
single \emph{constraint attribute} satisfying a user-specified
constraint. The algorithm design and performance evaluation for this
special case (with $m = 1$) generalize readily to the multi-criterion
problem; see Section~\ref{sec:general_framework}. An example that fits
this template that is of significant interest in the finance
community: the optimization of the expected return, subject to a risk
constraint. Here, the objective attribute is the mean of an arm
distribution, while the constraint attribute measuring risk could be
Conditional Value at Risk (CVaR).

For this problem, we propose an algorithm, called Constrained
Lower Confidence Bound (\textsc{Con-LCB}), that guarantees logarithmic
regret, i.e., the average number of plays of all non-optimal arms
(including those that violate the constraint) is at most
logarithmic in the horizon. If \textsc{Con-LCB} is presented with an
\emph{infeasbile} instance, i.e., an instance where all arms violate
the specified risk constraint, the algorithm in effect relaxes this
constraint just enough to make at least one arm compliant. Another
feature of \textsc{Con-LCB} is that at the end of the horizon, it outputs a
boolean flag that correctly identifies with high probability, whether
or not the given instance was feasible. 

Finally, we establish fundamental lower bounds on the performance of
any algorithm on this constrained regret minimization problem. Our
results demonstrate a fundamental tradeoff between regret minimization
and feasibility identification, similar to the well-known tradeoff
between regret minimization and best arm identification in the
classical (unconstrained) MAB problem \cite{bubeck2009}.

The remainder of this paper is organized as follows. A brief survey of
related literature is provided below.  In
Section~\ref{sec:formulation}, we introduce some preliminaries and
formulate the constrained mean minimization problem. We present
our algorithm (\textsc{Con-LCB}) and its performance guarantees in
Section~\ref{sec:algorithms}. Information-theoretic lower bounds on
performance are discussed in Section~\ref{sec:lower_bounds}. Finally,
the general formulation for multi-criterion MABs is introduced in
Section~\ref{sec:general_framework}.
All proofs are deferred to an appendix, which is part of the
`supplementary materials' document uploaded separately.

\noindent {\bf Related literature:}
The literature related to multi-armed bandit problems is quite
large. We refer the reader to \cite{bubeck2012, lattimore2018} for a
comprehensive review. Here, we restrict ourselves to papers that
consider (i) multi-objective MAB problems with vector rewards, and
(ii) risk-aware arm selection.

For multi-objective MAB problems with vector rewards, different
notions of optimality are considered.  For example, \cite{drugan2013,
  yahyaa2015} consider the notion of Pareto optimality.
In these papers, all dimensions are considered equally important and
the aim is to play all Pareto-optimal arms an equal number of
times. Another important notion of optimality is lexicographic
optimality (see \cite{ehrgott2005}). Here there is an order of
importance among different dimensions.  \cite{tekin2018, tekin2019}
consider the notion of lexicographic optimality for contextual bandit
problems.  In this line of work, the goal is to obtain higher reward
in an important dimension and for tie-breaking, use rewards obtained
in dimensions of lower importance.

Turning now to the literature on risk-aware arm selection,
\cite{sani2012, galichet2013, vakili2016, david2016, bhat2019, 
prashanth2019, kagrecha2019} consider the problem of optimizing 
a risk metric alone or consider a linear combination of mean and 
a risk metric. \cite{zimin2014} looks at the learnability of general
functions of mean and variance, and \cite{maillard2013} proposes an
optimization of the logarithm of moment generating function as a risk
measure in a regret minimization framework. \cite{cassel2018} look
at path dependent regret and provides a general approach to study many risk
metrics.

\ignore{None of the above papers considers the \emph{constrained} MAB problem,
which frames the optimal arm as the solution of a constrained
optimization problem. The only papers we are aware of that take this
approach in MAB setting are \cite{david2018} and
\cite{chang2020}. Both these papers consider a single constraint on
arm selection; \cite{david2018} considers a constraint on VaR, and
\cite{chang2020} considers an average cost constraint (each arm has a
cost distribution that is independent of its reward
distribution). While the present paper focuses on regret minimization,
\cite{david2018} works in the fixed confidence setting, and
\cite{chang2020} analyses the asymptotic probability of playing
optimal arms as the performance criterion. Aside from these
differences, the framework in this paper is more general, since we
allow for multiple constraints with arbitrary dependencies. Moreover,
\cite{david2018} and \cite{chang2020} implicitly assume that the
instance presented is feasible, whereas we address the issue of
encountering an infeasible instance.}

\revision{
None of the above papers considers the \emph{constrained} MAB problem,
which frames the optimal arm as the solution of a constrained
optimization problem. There has been some recent work for the constrained 
MAB problem. A constrained linear bandit setting is considered in 
\cite{pacchiano2021} under the assumption that there is at least one
arm which satisfies the constraints. The papers \cite{amani2019, moradipari2019} 
consider the problem of maximizing the reward subject to satisfying 
a linear constraint with high probability. Constrained setting was
also considered in \cite{david2018} and \cite{chang2020}. 
Both these papers consider a single constraint on
arm selection; \cite{david2018} considers a constraint on VaR, and
\cite{chang2020} considers an average cost constraint (each arm has a
cost distribution that is independent of its reward
distribution). All the papers above implicitly assume that the
instance presented is feasible, whereas we address the issue of
encountering an infeasible instance.
}


\section{Problem formulation}
\label{sec:formulation}
In this section, we describe the formulation of our constrained
stochastic MAB problem. To keep the exposition simple, we consider a
single constraint (i.e., $m = 1$) for now; the general formulation
with multiple constraints (i.e., $m \geq 1$) is discussed in
Section~\ref{sec:general_framework}. Informally, under the regret
minimization objective considered here, the goal is to play, as often
as possible, the arm that optimizes the objective, subject to a
constraint on another attribute.

Formally, consider a multi-armed bandit problem with $K$ arms, labeled
$1,2,\cdots,K.$ Each arm is associated with a (possibly
multi-dimensional) probability distribution, with $\nu(k)$ denoting
the (joint) distribution corresponding to arm~$k \in [K]$\footnote{For a
positive integer~$n,$ we denote $[n] := \{1,2,\cdots,n\}.$}.  Suppose that
$\nu(k) \in \mathcal{C},$ the space of possible arm distributions. The
objective and the constraint attributes are defined via functions
$g_0$ and $g_1$ respectively, mapping $\C$ to~$\R.$
Additionally, the user provides a threshold~$\tau \in \R$ which
specifies an upper bound on the attribute~$g_1.$ An \emph{instance} of
the constrainted MAB problem is defined by $(\nu,\tau),$ where $\nu =
(\nu(k),\ 1 \leq k \leq K).$ The arms which satisfy the constraint
$g_1(\nu(\cdot)) \leq \tau$ are called {\it feasible arms}; the set of
feasible arms is denoted by $\mathcal{K}(\nu).$ The instance
$(\nu,\tau)$ is said to be feasible if
$\mathcal{K}(\nu) \ne \emptyset,$ and is said to be infeasible if
$\mathcal{K}(\nu) = \emptyset.$

\begin{figure}
     \centering
     \begin{subfigure}[b]{0.45\textwidth}
         \centering
         \includegraphics[width=\textwidth]{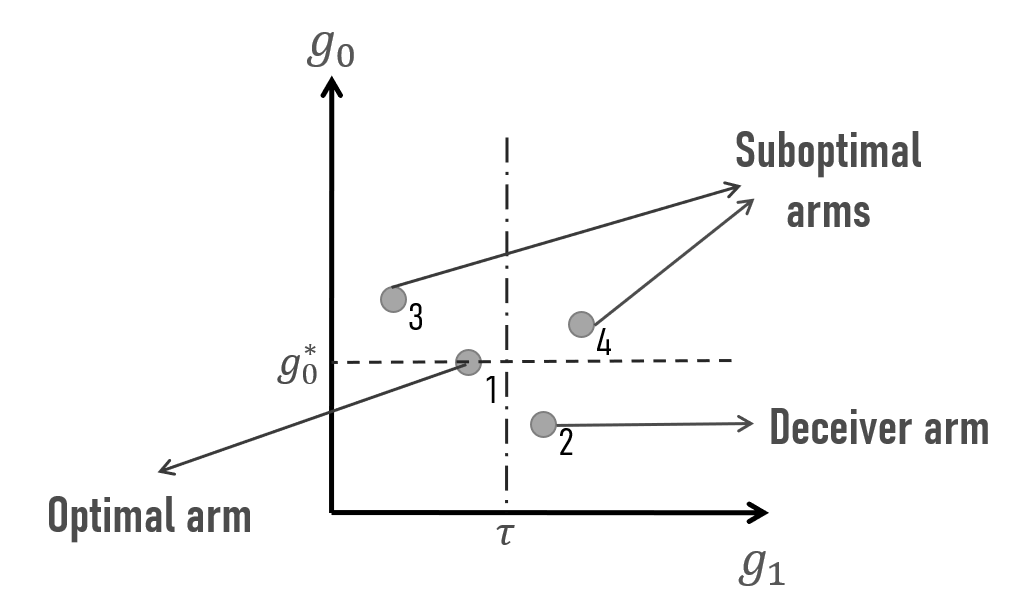}
         \caption{Feasible instance}
         \label{fig:feas_ins}
     \end{subfigure}
     \hfill
     \begin{subfigure}[b]{0.45\textwidth}
         \centering
         \includegraphics[width=\textwidth]{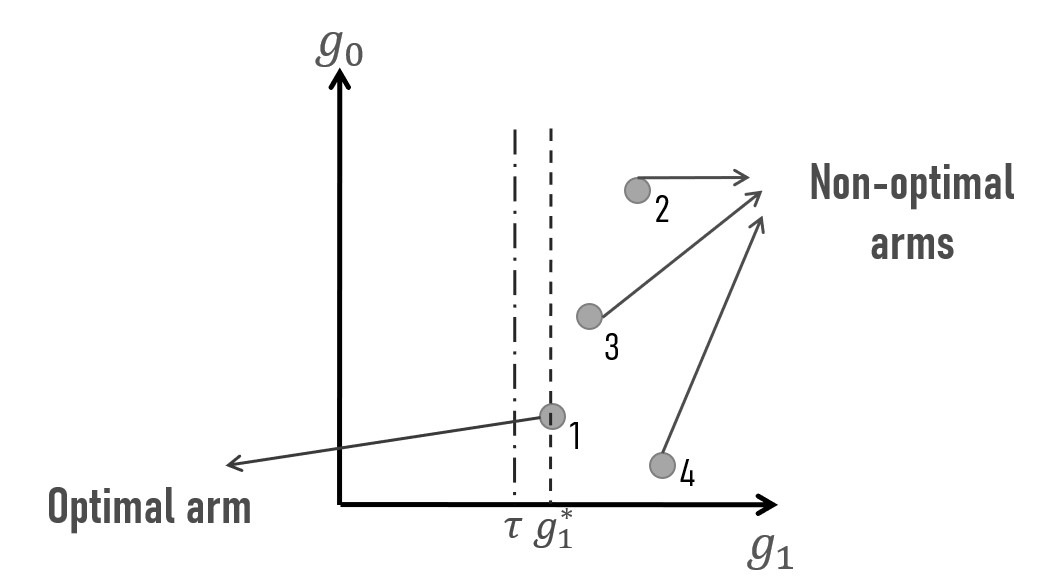}
         \caption{Infeasible instance}
         \label{fig:infeas_ins}
     \end{subfigure}
\end{figure}

Consider first a feasible instance. An optimal arm in this case is
defined as an arm that minimizes $g_0(\nu(\cdot)),$ subject to the
constraint $g_1(\nu(\cdot)) \leq \tau$. Denote the optimal value as
$g_0^* = \min_{k \in \mathcal{K}(\nu)} g_0(\nu(k)).$ Arms which have
$g_0(\nu(\cdot))$ larger than $g_0^*$ (whether or not feasible) are
referred to as \emph{suboptimal} arms. Note that there can also exist
infeasible arms with a smaller objective $g_0$ than $g_0^*.$ We refer
to such arms as {\it deceiver arms}; the set of deceiver arms is
denoted by $\mathcal{K}^d(\nu),$ where $\mathcal{K}^d(\nu) = \{k \in
[K] : g_1(\nu(k)) > \tau,~ g_0(\nu(k))
\leq g_0^*\}.$ For a suboptimal arm $k,$ the suboptimality gap is
defined by $\Delta(k) := g_0(k) - g_0^* > 0.$ (Note that the
suboptimality gap is also defined for infeasible, non-deceiver
arms). Finally, for an infeasible arm~$k,$ the infeasibility gap is
defined as $\Delta_{\tau}(k) = g_1(k) - \tau > 0.$
Figure~\ref{fig:feas_ins} provides a visual representation of a
feasbile instance.

Next, consider an infeasible instance. The optimal arm in this case is
defined in as the one with the smallest value of $g_1(\nu(\cdot)).$
Let the optimal value be denoted by $g_1^* = \min_{k \in [K]}
g_{1}(\nu(k)) > \tau.$ This is equivalent to requiring that if the
algorithm is faced with an infeasible instance, it must 'relax' the
constraint just enough, until at least one arm satisfies the
constraint. The \emph{constraint gap} for an arm~$k$ that is not
optimal is defined as $\Delta_{\text{con}}(k) = g_1(\nu(k)) - g_1^* >
0.$ Figure~\ref{fig:infeas_ins} provides a visual representation of an
infeasbile instance.

For any (feasible or infeasible) instance $(\nu,\tau),$ let the set of
optimal arms be denoted by $\mathcal{K}^*(\nu).$ The total number of
pulls (or horizon) is denoted by $T.$ For an algorithm (a.k.a.,
policy)~$\pi$, the number of pulls of an arm~$k$ over the first~$t$
pulls, for $t \in [T],$ is denoted by $N^{\pi}_{k}(t),$ though we
often suppress the dependence on $\pi$ for simplicity.

\noindent{\bf Consistency:} We now define the notion of
\emph{consistency} of an algorithm in this setting. A policy~$\pi$ is
said to be {\it consistent} over a class of distributions
$\mathcal{C},$ given a pre-specified constraint threshold $\tau,$ if
for all instances $(\nu,\tau)$ such that $\nu \in \mathcal{C}^K,$ it
holds that $\Exp{N_{k}(T)} = o(T^a)$ for all $a > 0$ and for all
(non-optimal) arms in $[K] \setminus \mathcal{K}^*(\nu).$ This
definition is in line with the definition of consistency in the
classical unconstrained regret minimization setting (see
\cite{lai1985}).

\noindent{\bf Regret:} The formal definition of regret in the present
setting is as follows. For a feasible instance, there are two types of
regret: \emph{suboptimality regret}
$$\Rsub_T := \sum_{k \in \mathcal{K}(\nu) \setminus \mathcal{K}^*(\nu)} 
\Delta(k) \Exp{N_{k}(T)},$$ which is the regret caused due to the
sampling of feasible, suboptimal arms, and \emph{infeasibility regret}
$$\Rinf_T
:= \sum_{k \in \mathcal{K}(\nu)^c} \Delta_\tau(i) \Exp{N_{k}(T)},$$
which is the regret caused due to the sampling of infeasible arms.
For an infeasible instance, regret is caused by playing arms that are
farther from the constraint boundary than the optimal arm. In this
case, we define the \emph{constraint regret} as
$$\Rcon_T := \sum_{k \in [K] \setminus\mathcal{K}^*(\nu)} 
\Delta_{\text{con}}(k) \Exp{N_{k}(T)}.$$ (Note that our analysis
essentially provides upper and lower bounds on $\Exp{N_{k}(T)}$ for
all arms in $[K] \setminus\mathcal{K}^*(\nu).$ So alternative
definitions of regret, involving linear combinations of the expected
number of pulls of non-optimal arms, are also supported by our
analysis.)

\noindent{\bf Infeasibility identification:} Next, we introduce an
optional boolean flag called \texttt{feasibility\_flag}, that the
policy may set at the end of $T$ plays, to indicate post facto whether
it considered the instance as feasible (by setting the flag as
\texttt{true}) or infeasible (by setting the flag as
\texttt{false}). For the algorithms proposed in this paper, we provide
bounds on the probability that this flag erroneously flags a feasible
instance as infeasible and vice-versa. We also provide fundamental
lower bounds on the probability of error under any consistent policy.

\noindent{\bf Concentration inequalities on attribute estimators:}
Natually, MAB algorithms must estimate the attributes $g_0$ and $g_1$
corresponding to each arm using the data samples obtained by pulling
that arm. We make the following assumption on concentration properties
of these estimators. Suppose that for $i \in \{0,1\},$ and
distribution $F \in \mathcal{C},$ there exists an estimator
$\hat{g}_{i,n}(F)$ of $g_i(F)$ using $n$ i.i.d. samples from~$F,$
satisfying the following concentration inequality: There exists~$a_i >
0$ such that for all $\Delta > 0,$
\begin{equation}
  \label{eq:gen_conc_ineq_m1}
  \prob{\left|\hat{g}_{i,n}(F) - g_i(F) \right| 
  \geq \Delta} \leq 2 \exp(-a_i n \Delta^2).
\end{equation}
Concentration inequalities of this form are available in a broad
variety of settings. For example, if $g_i(F) = \Exp{h_i(X)},$ where
$X$ is a random vector distributed as~$F,$ then a concentration
inequality of the form~\eqref{eq:gen_conc_ineq_m1} is readily obtained
if $h_i(X)$ is bounded (using the Hoeffding inequality), or
$\sigma$-subGaussian. Similarly, if $h_i$ is Lipschitz and~$X$ is a
subGaussian random vector, concentration bounds of the
form~\eqref{eq:gen_conc_ineq_m1} can be obtained by invoking the
results in \cite{kontorovich2014}. Several examples where risk
measures can be concentrated in this manner are provided
in~\cite{cassel2018}. Also, note that the specific form
\eqref{eq:gen_conc_ineq_m1} of the concentration inequality is only
assumed to simplify the description of our algorithms. Alternative
forms of concentration guarantees (such as those known for the means
of subexponential or heavy-tailed distributions) can also be supported
by our algorithmic framework via trivial modifications to the
confidence bounds.

\section{Constrained regret minimization}
\label{sec:algorithms}
In this section, we present an algorithm for constrained regret
minimization, and provide performance guarantees for the same,
assuming that we have estimators for each attribute that satisfy the
concentration inequality~\eqref{eq:gen_conc_ineq_m1}.

The algorithm, which we refer to as \emph{constrained lower confidence
  bound} (\textsc{Con-LCB}) algorithm, is based on the well-known
  principle of \emph{optimism under uncertainty}. \textsc{Con-LCB}
  uses lower confidence bounds (LCBs) on attribute $g_1$ of each arm
  to maintain a set of \emph{plausibly feasible} arms, and uses LCBs
  for attribute $g_0$ of the arms in this set to select the arm to be
  played. Note that LCBs are used for attribute $g_0$ because we are
  dealing with \emph{minimization} of $g_0.$
If, at some instant, the set of plausibly feasible arms maintained by
\textsc{Con-LCB} becomes empty, the algorithm turns conservative and plays the
arm which violates the constraint least, i.e., the one with the smallest LCB
on $g_1$. Finally, at the end of~$T$ rounds, \textsc{Con-LCB} sets the feasiblity
flag as \texttt{true} if the set of plausibly feasible arms is found
to be non-empty, and \texttt{false} otherwise.

\begin{algorithm}
  \caption{\textsc{Con-LCB}}
  \label{alg:con_lcb}
\begin{algorithmic}
  \Procedure{\textsc{Con-LCB}}{$T, K, \tau$}
  \State $\text{Play each arm once}$
  \For{$t=K+1,\cdots,T$}
    \State Set $\hat{\calk}_{t} = \left\{ k: \hat{g}_{1, N_{k}(t-1)}(k) - \sqrt{\frac{\log(2T^2)}{a_1 N_{k}(t-1)}} \leq \tau \right\}$
    \If{$\hat{\calk}_t \neq \varnothing$}
    \State Set $\small \mathrm{L}_0(k) = \hat{g}_{0,N_{k}(t-1)}(k) - \sqrt{\frac{\log(2T^2)}{a_0 N_{k}(t-1)}}$ 
    \State $\text{Play arm }k_{t}^{\dagger} \in \argmin_{k \in \hat{\calk}_t} \mathrm{L}_0(k)$
    \Else
    \State Set $\small \mathrm{L}_1(k) = \hat{g}_{1,N_{k}(t-1)}(k) - \sqrt{\frac{\log(2T^2)}{a_1 N_{k}(t-1)}}$ 
    \State $\text{Play arm }k_{t}^{\dagger} \in \argmin_{k \in [K]}  \mathrm{L}_1(k)$
    \EndIf
  \EndFor
  \If{$\hat{\calk}_{T} \neq \varnothing$}
    \State $\text{Set } \texttt{feasibility\_flag} = \texttt{true}$
  \Else
    \State $\text{Set } \texttt{feasibility\_flag} = \texttt{false}$ 
  \EndIf
  \EndProcedure
\end{algorithmic}
\end{algorithm} 

The remainder of this section is devoted to performance guatantees for
\textsc{Con-LCB}. We consider feasible and infeasible instances separately.

\begin{theorem}
\label{thm:con_lcb_feasible1}
Consider a feasible instance. Under \textsc{Con-LCB}, the expected
number of pulls of a feasible but suboptimal arm $k$ (i.e., satisfying
$g_0(\nu(k)) > g_0^*$ and $g_1(\nu(k)) \leq \tau$), is bounded by
\begin{align*}
\Exp{N_{k}(T)} \leq \frac{4 \log(2T^2)}{a_0 \Delta^2(k)} + 5.    
\end{align*}  
The expected number of pulls of a deceiver arm $k$ (i.e., satisfying
$g_0(\nu(k)) \leq g_0^*$ and $g_1(\nu(k)) > \tau$ is bounded by
\begin{align*}
\Exp{N_k(T)} \leq \left(\frac{4 \log(2T^2)}{a_1 [\Delta_{\tau}(k)]^2}\right) + 2.
\end{align*}
The expected number of pulls of an arm $k$ which is an infeasible
non-deceiver (i.e., satistying $g_0(\nu(k)) > g_0^*$ and $g_1(\nu(k))
> \tau$) is bounded by
\begin{align*}
\Exp{N_k(T)} \leq \min \left(\frac{4 \log(2T^2)}{a_0 \Delta^2(k)}, 
\frac{4 \log(2T^2)}{a_1 [\Delta_{\tau}(k)]^2}  \right) + 5.
\end{align*}
The probability of incorrectly setting the \emph{\texttt{feasibility\_flag}}
is upper bounded by
\begin{align*}
  \prob{\emph{\texttt{feasibility\_flag = false}}} \leq \frac{1}{T}.
\end{align*}
\end{theorem}

The main takeaways from Theorem~\ref{thm:con_lcb_feasible1} are as
follows.

$\bullet$~The upper bound on the expected number of pulls
for feasible, suboptimal arms is logarithmic in the horizon~$T,$ and
inversely proportional to the square of the suboptimality gap. This is
similar to bounds known for classical (unconstrained) MABs.

$\bullet$~For deceiver arms, the upper bound on the
expected number of pulls, also logarithmic in $T,$ is inversely
proportional to the square of the feasiblity gap. This is similar to
the bound one would obtain in a pure~$g_1$ minimization problem for a
hypothetical instance consisting of all the originally infeasible
arms, and a single hypothetical (optimal) arm having $g_1$ equal to
$\tau.$

$\bullet$~The bound on the expected number of pulls of
non-deceiver, infeasible arms involves a minimum of the dominant terms
in the above two cases. Intuitively, this is because these arms can
disambiguated in two ways: via the suboptimality gap, and the
feasibility~gap.

$\bullet$~The probability that the feasiblity flag
incorrectly identifies the instance as infeasible is upper bounded by
a power law in the horizon~$T.$ Note that the specific form of the
probability of mis-identification bound is not fundamental; a small
modification in the algorithm would make this bound a faster decaying
power law at the expense of a multiplicative bump in regret. However,
that this probability is not \emph{much} smaller (for example,
exponentially decaying in the horizon~$T$) is a consequence of an
inherent tension between regret minimization and feasiblity
identification.
A similar tension is known to exist between regret minimization and
best arm identification in the unconstrained setting; see
\cite{bubeck2009}). We provide lower bounds on the probability that
any consistent algorithm makes a mistake in feasibility
identification in Section~\ref{sec:lower_bounds}.

Finally, we note that the suboptimality regret, as well as the
infeasibility regret are logarithmic in the horizon~$T.$ Indeed, the
suboptimality regret for a feasible instance is bounded as
\begin{equation*}
\Rsub_T \leq \sum_{k \in \mathcal{K}(\nu) \setminus \mathcal{K}^*(\nu)} 
\left(\frac{4 \log(2T^2)}{a_0\Delta(k)} + 5 \Delta(k) \right),
\end{equation*}
which is similar to
regret bounds in the classical unconstrained setting. To express the
infeasibility regret compactly, let us interpret $\Delta(k)$ as 0 (and
$1/\Delta(k)$ as $\infty$) for deceiver arms. With this notation, the
infeasibility regret of a feasible instance is bounded as
\begin{equation*}
\Rinf_T \leq \sum_{k \in \mathcal{K}(\nu)^c} 5\Delta_\tau(k) + 
\min \left(\frac{4\Delta_\tau(k) \log(2T^2)}{a_0\Delta^2(k)}, \frac{4\log(2T^2)}{a_1 \Delta_\tau(k)} \right).
\end{equation*}

Next, we move on to characterizing the performance of \textsc{Con-LCB} over
infeasible instances.
\begin{theorem}
\label{thm:con_lcb_infeasible1}
Consider an infeasible instance. Under \textsc{Con-LCB}, the expected
number of pulls of a non-optimal arm~$k$ is bounded by
\begin{equation*}
  \Exp{N_{k}(T)} \leq \left(\frac{4 \log(2T^2)}{a_1 [\Delta_{\rm {con}}(k)]^2}\right) + K + 2.
\end{equation*}
Moreover, the probability that the algorithm incorrectly flags the
instance as feasible is bounded as
$\prob{\emph{\texttt{feasibility\_flag}} = \emph{\texttt{true}}} \leq
\frac{K}{T} \text{ for } T>T^*(\nu),$ where $T^*(\nu)$ is an
instance-dependent constant.
\end{theorem}
For an infeasible instance, the upper bound on the expected number of
pulls of a non-optimal arm, logarithmic in the horizon, and inversely
proportional to the the square of the constraint gaps $\Delta_{\rm
{con}}(k),$ is structurally similar to the bound one would obtain in
pure~$g_1$ minimization problem on the same instance. However, note
that when faced with an infeasible instance, \textsc{Con-LCB} would
only start playing the optimal arm regularly after \emph{all}~$K$ arms
appear to be infeasible; this explains the appearance of~$K$ in our
bounds.

Here, the constraint regret is bounded as
\begin{equation*}
\Rcon_T := \sum_{k \in [K] \setminus\mathcal{K}^*(\nu)} (K+2) \Delta_{\rm {con}}(k) 
+ \frac{4 \log(2T^2)}{a_1 \Delta_{\rm {con}}(k)} \end{equation*}
 Finally, as before, the probability that the feasibility flag wrongly
 identifies the infeasible instance as feasible decays as a power law
 in the horizon for $T > T^*(\nu);$ the threshold~$T^*$ accounts for
 the time it takes for the algorithm to `detect' that the instance is
 infeasibile with high probability.


\section{Information theoretic lower bounds}
\label{sec:lower_bounds}
In this section, we establish fundamental limits on the performance of
algorithms for constrained regret minimization. First, we show that
the regret bounds obtained for \textsc{Con-LCB} are asymptotically
tight upto universal multiplicative constants (on a class of Gaussian
bandit instances).
We then prove a lower bound on the probability that any consistent
algorithm misidentifies a feasible instance as infeasible or
vice-versa. This result illustrates an inherent tension between regret
minimization and feasibility identification---consistent algorithms
(recall that consistency means regret is $o(T^a)$ for all $a > 0$)
cannot have a misidentification probability that decays exponentially
in the horizon, and algorithms that enjoy a mid-identification
probability that decays exponentially in the horizon cannot be
consistent.

To state our information theoretic lower bound on regret, suppose that
the class of arm distributions $\mathcal{G}$ is the class of
2-dimensional Gaussian distributions with covariance matrix $\Sigma
= \mathrm{diag}(\nicefrac{1}{2 a_0}, \nicefrac{1}{2 a_1}).$ Let
attribute $g_0$ be the mean of the first dimension, and $g_1$ be the
mean of the second dimension. For the assumed covariance matrix
structure, the standard empirical mean estimators for $g_0$ and $g_1$
satisfy the concentration properties stated
in~\eqref{eq:gen_conc_ineq_m1}.
\begin{theorem}
\label{thm:lower_bound_gaussian}
Let $\pi$ be any consistent policy over the class of distributions
$\mathcal{G}$ given the threshold $\tau \in \mathbb{R}.$ Consider a
feasible instance $(\nu^f,\tau),$ where $\nu^f \in \mathcal{G}^K.$ For
any feasible but suboptimal arm $k,$
\begin{equation*}
\liminf_{T \to \infty} \frac{\Exp{N^{\pi}_{k}(T)}}{\log(T)} \geq \frac{1}{a_0 \Delta^2(k)}.
\end{equation*}
For any deceiver arm $k,$
\begin{align*}
\liminf_{T \to \infty} \frac{\Exp{N^{\pi}_{k}(T)}}{\log(T)} \geq \frac{1}{a_1 [\Delta_\tau(k)]^2}.
\end{align*}
Finally, for any infeasible non-deceiver arm $k,$ 
\begin{align*}
\liminf_{T \to \infty} \frac{\Exp{N^{\pi}_{k}(T)}}{\log(T)} \geq  
\frac{1}{2}\min\left(\frac{1}{a_0 \Delta^2(k)}, \frac{1}{a_1 [\Delta_\tau(k)]^2} \right).
\end{align*}
Similarly, for an infeasible instance $(\nu^i,\tau),$ such that
$\nu^i \in \mathcal{G}^K,$ for any non-optimal arm $k,$
\begin{align*}
\liminf_{T \to \infty} \frac{\Exp{N^{\pi}_{k}(T)}}{\log(T)} \geq  
\frac{1}{a_1 [\Delta_\text{con}(k)]^2}.
\end{align*}
\end{theorem} 
The proof of Theorem~\ref{thm:lower_bound_gaussian} can be found in
Appendix~\ref{app:lower_bound}. Comparing the lower bounds in
Theorem~\ref{thm:lower_bound_gaussian} with the upper bounds
for \textsc{Con-LCB} in Theorems~\ref{thm:con_lcb_feasible1}
and~\ref{thm:con_lcb_infeasible1}, we conclude that \textsc{Con-LCB}
is asymptotically optimal on regret, up to universal multiplicative
constants.\footnote{Information theoretic lower bounds on the expected
number of pulls of non-optimal arms can be derived (in terms of a
certain optimization over KL divergences) for any general class of arm
distributions; see Appendix~\ref{sec:gaussian_pulls}. We have
specialized the statement of Theorem~\ref{thm:lower_bound_gaussian} to
a class of Gaussian bandit instances to enable an easy comparison with
our upper bounds for \textsc{Con-LCB}.}


Next, we address the fundamental tradeoff between regret minimization
and feasibility identification.

\begin{theorem}
  \label{thm:prob_incorrect_flag}
  Consider a space of arm distributions~$\mathcal{C},$ and a 
  threshold~$\tau$ such that $\mathcal{C}$ contains both feasible as
  well as infeasible arm distributions. There exists a feasible
  instance~$(\nu,\tau)$ and an infeasible instance~$(\nu', \tau)$ such
  that for any policy~$\pi$ that is consistent over~$\mathcal{C},$
\begin{align*}
  \limsup_{T \to \infty} -\frac{1}{T} &\log\Big(\probnu{\nu}{\emph{\texttt{feasibility\_flag}}= \emph{\texttt{false}}} \\
  &\quad + \probnu{\nu'}{\emph{\texttt{feasibility\_flag}}= \emph{\texttt{true}}}\Big) \leq 0.
\end{align*}
\end{theorem} 
Theorem~\ref{thm:prob_incorrect_flag} states that for any consistent
algorithm, the probability that~$(\nu,\tau)$ get misidentified as
infeasible, and the probability that~$(\nu',\tau)$ get misidentified
as feasible, cannot both decay exponentially in the horizon. This is
of course consistent with the power-law probability of
misidentification under \textsc{Con-LCB}. In other words,
slower-than-exponential decay of the probability of feasibility
misidentification with respect to the horizon is an unavoidable
consequence of the exploration-exploitation interplay in regret
minimization. A similar tension between regret minimization and best
arm identification was noted for the unconstrained MABs by
\cite{bubeck2009}.

\section{General framework for constrained MABs}
\label{sec:general_framework}

In this section, we provide a general formulation for constrained
stochastic MABs with multiple criteria. We allow each arm to be
associated with a $D$-dimensional probability distibution, the goal
being to optimize one (dominant) attribute associated with this
distribution, subject to constraints on~$m$ others. The algorithm
design and performance evaluation performed in
Sections~\ref{sec:formulation}--\ref{sec:lower_bounds} for the special
case of $m=1$ extend naturally to this
general formulation, which can in turn be applied in a variety of
application scenarios. We illustrate a few here.

$\bullet$~For clinical trials, with the arms corresponding
to various treatment protocols, the dominant attribute might, for
example, correspond to the success/recovery probability, whereas the
constraints might capture recovery time, severity of side effects,
etc.
  
$\bullet$~For product/service rating, where the arms correspond to
various service providers, the dominant attribute might correspond to
product quality, with constraints capturing reliability, pricing,
customer service, etc.

$\bullet$~In wireless networks, the arms might correspond to various
access networks or channels, with, for example, the dominant attribute
corresponding to throughput, and constraints capturing delay, energy
efficiency, etc.

\noindent{\bf Formulation:}
Consider a set of $K$ arms, each associated with a $D$-dimensional
probability distribution, with $\nu(k)$ denoting the joint distribution
corresponding to arm~$k \in [K].$ Suppose that
$\nu(k) \in \mathcal{C},$ the space of possible arm distributions. The
objective and the constraints are defined by functions
$g_0, g_1, \cdots, g_m.$ Specifically, the optimal arm is defined as
that arm~$k$ that minimizes~$g_0(\nu(k)),$ subject to the constraints
$\{g_i(\nu(k)) \leq \tau_i\}_{i=1}^m$ when the instance is feasible
(i.e., at least one arm exists that satisfies the above constraints).

If the instance is infeasible, the optimal arm is defined via a ranked
list of the constraints, that orders them by `importance'. Without
loss of generality assume that the order of importance increases from
$(g_1, \tau_1)$ to $(g_m, \tau_m).$ The idea is to relax the
constraints one-by-one, starting with the least important, until a
compliant arm is found. Formally, for a given infeasible
instance~$\nu$, for $2 \leq i \leq m,$ let $\mathcal{K}_i(\nu)$ denote
the set of arms that satisfy the constraints
$\{g_j(\nu(k)) \leq \tau_j\}_{j=i}^m.$ Let us also define
$\mathcal{K}_{m+1}(\nu):= [K].$ Now, let
$i^*(\nu) = \min \{k \in [m]:\ \mathcal{K}_{k+1}(\nu) \ne \phi\}.$
Here,~$i^*(\nu)$ is the fewest number of constraints one must relax,
in order of increasing importance, in order to have at least one
compliant arm. An optimal arm is then defined to be
$\argmin \{g_{i^*(\nu)}(\nu(k)):\ k \in \mathcal{K}_{i^*(\nu)+1}(\nu)\}.$

Similar to the case where $m=1,$ we make the following assumption 
to simplify algorithm design. Suppose
that for $0 \leq i \leq m,$ and $\overline{\nu} \in \mathcal{C},$ 
there exist an estimator $\hat{g}_{i,n}(\overline{\nu})$ for 
$g_i(\overline{\nu})$ using $n$ i.i.d. samples from~$\overline{\nu},$ 
satisfying the following concentration inequality: There
exists~$a_i > 0$ such that for all $\Delta > 0,$
\ignore{Suppose in the underlying instance, none of the arms satisfy
  constraint $(g_j, \tau_j),$ then the constraints
  ${(g_i, \tau_i)}_{i=1}^{j}$ are revoked and the goal becomes to
  maximize the pulls of an arm which minimizes $\Exp{f(\Xbf)}$ subject
  to constraints ${(g_i, \tau_i)}_{i=j+1}^{m}$ when $j<m.$ If the most
  important constraint is not satisfied by any of the arms, i.e.,
  $j=m,$ then all constraints are revoked.}
\ignore{Assume that there exist estimators
  $\overline{g}_0(X), \cdots, \overline{g}_m(X)$ for
  $\Exp{g_0(X)}, \cdots, \Exp{g_m(X)}$ such that we have a
  concentration inequality of the following form:} 
\begin{equation}
  \label{eq:gen_conc_ineq}
  \prob{\left|\hat{g}_{i,n}(\overline{\nu}) - g_i(\overline{\nu}) \right| 
  \geq \Delta} \leq 2 \exp(-a_i n \Delta^2).
\end{equation}
Note that this is simply an extension of our assumption~\eqref{eq:gen_conc_ineq_m1}. 


\ignore{
\noindent{\bf Algorithm and performance guarantees:} For the above problem
formulation, we propose \textsc{Multi-Con-LCB}, a simple extension of 
the \textsc{Con-LCB} algorithm that we presented before for the case of $m=1.$
Due to space constraints, we defer the algorithm description to 
Appendix~\ref{app:gen_framework} but state the performance guarantees here
because of their similarity to Theorem~\ref{thm:con_lcb_feasible}. Before,
we state the theorem, note that $\Delta(k)$ denotes the suboptimality gap
like before and $\Delta_{i,\tau_i}(k)$ represents the infeasibility gap
for constraint $i$ and is similar to $\Delta_\tau(k)$ when $m=1.$   }

\revision{
\noindent{\bf Algorithm and performance guarantees:} For the above problem
formulation, we propose \textsc{Multi-Con-LCB}, a simple extension of 
the \textsc{Con-LCB} algorithm that we presented before for the case of $m=1.$
\textsc{Multi-Con-LCB} uses upper confidence bounds on constrained
attributes $\{g_i(\nu(k))\}_{i=1}^{m}$ for each arm $k$ to maintain a
set of plausibly feasible arms.  The set of plausibly feasible arms
for the constraint on function $g_i$ at time $t \in [T]$ will be
denoted by $\hat{\calk}_{i,t}^{\dagger}$ for all values of
$i \in [m].$ Using the sets of plausibly feasible arms for each
constraint, we construct the set of arms that plausibly lie in the set
$\calk(\nu)$ and this set is denoted by $\hat{\calk}_t$ for time instant
$t \in [T].$ Formally,
$\hat{\calk}_t = \cap_{i=1}^{m} \hat{\calk}_{i,t}^\dagger.$ As this
set might be empty, for $i \in \{2,\cdots,m\},$ let the estimate for
$\calk_i(\nu)$ be
$\hat{\calk}_{i,t} = \cap_{j=i}^{m} \hat{\calk}_{j,t}^\dagger.$ It is
also possible that the most important constraint is not satisfied,
therefore let $\hat{\calk}_{m+1, t} = [K].$ If the set $\hat{\calk}_t$
is not empty, then the algorithm uses lower confidence bounds (LCBs)
on $g_0$ to select the arm to be played. If the set $\hat{\calk}_t$
turns out to be empty, then the algorithm finds the smallest index
$\hat{i}^*$ such that the set $\hat{\calk}_{\hat{i}^*+1,t}$ is not
empty. The algorithm then plays the arm with lowest LCB on
$g_{\hat{i}^*}.$ Finally, at the end of $T$ rounds, \textsc{Multi-Con-LCB}
sets the feasibility flag as \texttt{true} if the set $\hat{\calk}_T$
is not empty and \texttt{false} otherwise. The details are presented
as Algorithm~\ref{alg:conlcb}.

\begin{algorithm}
  \caption{Multiple Constrained LCB}
  \label{alg:conlcb}
\begin{algorithmic}
  \Procedure{Multi-Con-LCB}{$T, K, \{\tau_{i} \}_{i=1}^{m}$}
  \State $\text{Play each arm once}$
  \For{$t=K+1,\cdots,T$}
    \For{$i=1,\cdots,m$}
    \State $\text{Set } \hat{\calk}_{i,t}^\dagger = \left\{ k: \hat{g}_{i, N_{k}(t-1)}(k) \leq 
    \tau_i + \sqrt{\frac{\log(2T^2)}{a_i N_{k}(t-1)}} \right\}$
    \EndFor
    \State $\text{Set } \hat{\calk}_t = \cap_{i=1}^{m} \hat{\calk}_{i,t}^\dagger$
    \If{$\hat{\calk}_t \neq \varnothing$}
      \State $k_{t+1}^{\dagger} \in \argmin_{k \in \hat{\calk}_t} 
      \hat{g}_{0,N_{k}(t-1)}(k) - \sqrt{\frac{\log(2T^2)}{a_0 N_{k}(t-1)}}$ 
      \State $\text{Play arm }k_{t+1}^{\dagger}$
    \Else
      \State $\hat{i}^* = \argmin_{i \in \{1,\cdots,m\}} \hat{\calk}_{i+1,t} \ne \varnothing$
      \State $k_{t+1}^{\dagger} \in \argmin_{k \in \hat{\calk}_{\hat{i}^*+1, t}} 
        \hat{g}_{\hat{i}^*,N_{k}(t-1)}(k) - \sqrt{\frac{\log(2T^2)}{a_{\hat{i}^*} N_{k}(t-1)}}$
      \State $\text{Play arm }k_{t+1}^{\dagger}$ 
    \EndIf
  \EndFor
  \If{$\hat{\calk}_{T+1} \neq \varnothing$}
    \State $\text{Set } \texttt{feasibility\_flag = true}$
  \Else
    \State $\text{Set } \texttt{feasibility\_flag = false}$
  \EndIf
  \EndProcedure
\end{algorithmic}
\end{algorithm}

The remainder of this section is devoted to performance guarantees for
\textsc{Multi-Con-LCB}. The suboptimality gap for an arm $k$ is given by
$\Delta(k) = \max(g_0(\nu_k) - g_0^*, 0).$ The infeasibility gap of an
arm $k$ for constraint $i$ is given by
$\Delta_{i, \tau_i}(k) = \max(g_i(\nu(k))-\tau_i, 0).$ We restrict our
attention to feasible instances here; infeasible instances can be
handled on similar lines as
Theorem~\ref{thm:con_lcb_infeasible1} for \textsc{Con-LCB}.}

\begin{theorem}
\label{thm:con_lcb_feasible}
Consider a feasible instance. Under \textsc{Multi-Con-LCB}, the expected
number of pulls of a feasible but suboptimal arm $k$ (i.e., satisfying
$g_0(\nu(k)) > g_0^*$ and $g_i(\nu(k)) \leq \tau_i$ for all
$i \in [m]$), is bounded by
\begin{align*}
\Exp{N_{k}(T)} \leq \frac{4 \log(2T^2)}{a_0 \Delta^2(k)} + 2m + 3.    
\end{align*}  
The expected number of pulls of a deceiver arm $k$ (i.e., satisfying $g_0(\nu(k)) \leq g_0(\nu(1))$ and there exists 
a constraint indexed by $j \in [m]$ such that $g_j(\nu(k)) > \tau_j$) is bounded by
\begin{align*}
\Exp{N_k(T)} \leq \min_{i \in [m]} \left(\frac{4 \log(2T^2)}{a_i [\Delta_{i, \tau_i}(k)]^2}\right) + m.
\end{align*}
The expected number of pulls of a an arm $k$ which is infeasible, but not a deceiver (i.e., satistying 
$g_0(\nu(k)) > g_0(\nu(1))$ and there exists a constraint indexed by $j \in [m]$ such that $g_j(\nu(k)) > \tau_j$) 
is bounded by
\begin{align*}
\Exp{N_k(T)} \leq \min \left(\frac{4 \log(2T^2)}{a_0 \Delta^2(k)}, 
\min_{i \in [m]} \left(\frac{4 \log(2T^2)}{a_i [\Delta_{i, \tau_i}(k)]^2}\right)  \right) 
+ 2m + 3.
\end{align*}
The probability of incorrectly setting the \emph{\texttt{feasibility\_flag}}
is upper bounded by
\begin{align*}
  \prob{\emph{\texttt{feasibility\_flag = false}}} \leq \frac{m}{T}.
\end{align*}
\end{theorem}
\ignore{
Note the similarities between Theorems~\ref{thm:con_lcb_feasible} and~\ref{thm:con_lcb_feasible1}. The guarantees for the expected 
number of pulls are worse only by an additive constant which is a function
of number of constraints. The probability of mis-identification gets 
multiplied by $m,$ i.e., the number of constraints. }
\revision{
We omit the proof of Theorem~\ref{thm:con_lcb_feasible} in the appendix
because it is very similar to the proof of Theorem~\ref{thm:con_lcb_feasible1}.
However, we state the key takeaways from Theorem~\ref{thm:con_lcb_feasible} below.

$\bullet$~Similar to Theorem~\ref{thm:con_lcb_feasible1}, the upper bound 
on the expected number of pulls of suboptimal arms is logarithmic in $T$ 
and is inversely proportional to the suboptimality gap squared. Moreover, 
the upper bound for a deceiver arm is also logarithmic in $T$ but there 
is a minimum over $m$ terms because the deceiver arm can be identified 
by any of the constraints that the arm does not satisfy. Similarly, the 
upper bound for a non-deceiver, suboptimal arm is also logarithmic in $T$
and there is a minimum over $m+1$ terms because the arm can be identified
as suboptimal, or infeasible for not satisfying one of the $m$ constraints.

$\bullet$~With an increase in the number of constraints to $m$, the upper
bounds on the expected number of pulls of non-optimal arms increase linearly in
$m$ and the probability of mis-identification also gets scaled by a 
factor of $m.$ The slight degradation in the guarantees is expected because
with more constraints, the optimal arm has to satisfy more conditions,
and it becomes harder to compare different arms.     
}


\section{Numerical experiments}
In this section, we present a simple experiment to show the numerical
performance of our algorithm \textsc{Con-LCB}.

We consider an instance with four arms and two criteria. Each arm is
associated with a (two-dimensional) multivariate Gaussian
distribution, different arms having the same covariance matrix
$\Sigma,$ but different mean vectors. The means corresponding to the
first dimension are 0.3, 0.4, 0.2, and 0.5 and the means corresponding to
the second dimension are 0.4, 0.4, 1.0, and 1.0. The covariance matrix
is given by $\Sigma = \begin{pmatrix}
  1 & 0.5 \\
  0.5 & 1
\end{pmatrix}.$

\begin{figure}[t]
\centering
\includegraphics[width=0.5\textwidth]{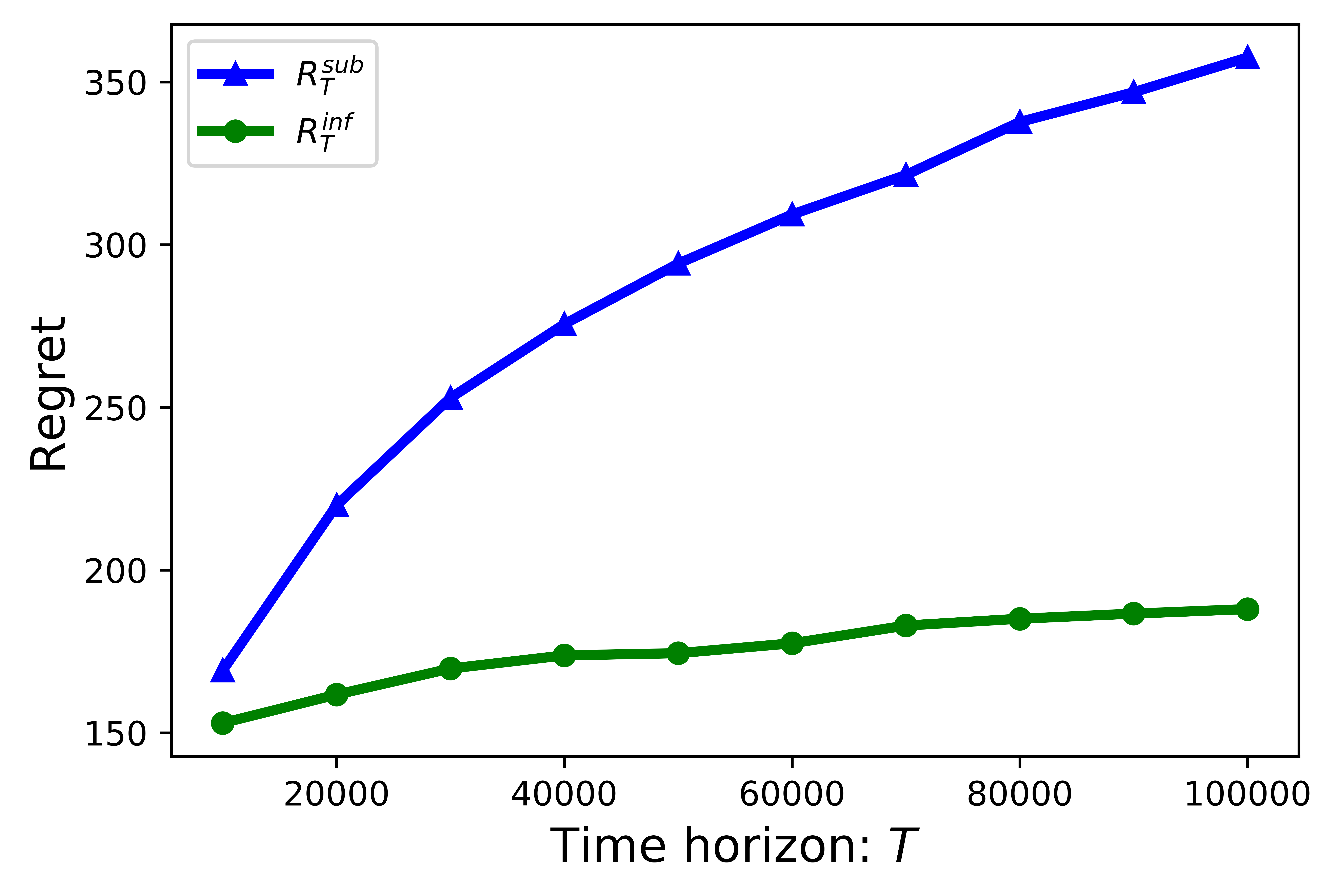}
\caption{Regret of Gaussian instance}
\label{fig:regret}
\end{figure}

The goal is to maximize the pulls of the arm with the minimum mean of
the first dimension, subject to a constraint on the mean of the
second. Specifically, we require that the mean of the second dimension
should be less than or equal to $\tau := 0.5.$ The instance is
summarized in Table~\ref{tab:means_remarks}. We use empirical averages
as the estimators for the means and standard confidence bounds
based on sub-Gaussianity assumption. We run the algorithm 1000
times for values of $T$ in $[10000, 100000].$ The suboptimality regret
$R_{T}^{\text{sub}}$ and the infeasibility regret $R_{T}^{\text{inf}}$
are plotted in Figure~\ref{fig:regret}. Clearly, the regrets grow 
sub-linearly with respect to the horizon. Also, in our experiments, 
the algorithm correctly detected the feasibility of the instance in 
all of the 1000 runs.

\begin{table}[h]
\caption{Gaussian instance}
\label{tab:means_remarks}
\begin{tabular}{cccl}
  \textbf{Arm} & \textbf{Mean 1} & \textbf{Mean 2} & \textbf{Remarks} \\ 
  1 & 0.3 & 0.4 & Optimal arm \\  
  2 & 0.4 & 0.4 & Feasible, suboptimal \\
  3 & 0.2 & 1.0 & Deceiver \\
  4 & 0.5 & 1.0 & Infeasible, suboptimal 
\end{tabular}
\end{table}


\begin{figure}[t]
\centering
\includegraphics[width=0.5\textwidth]{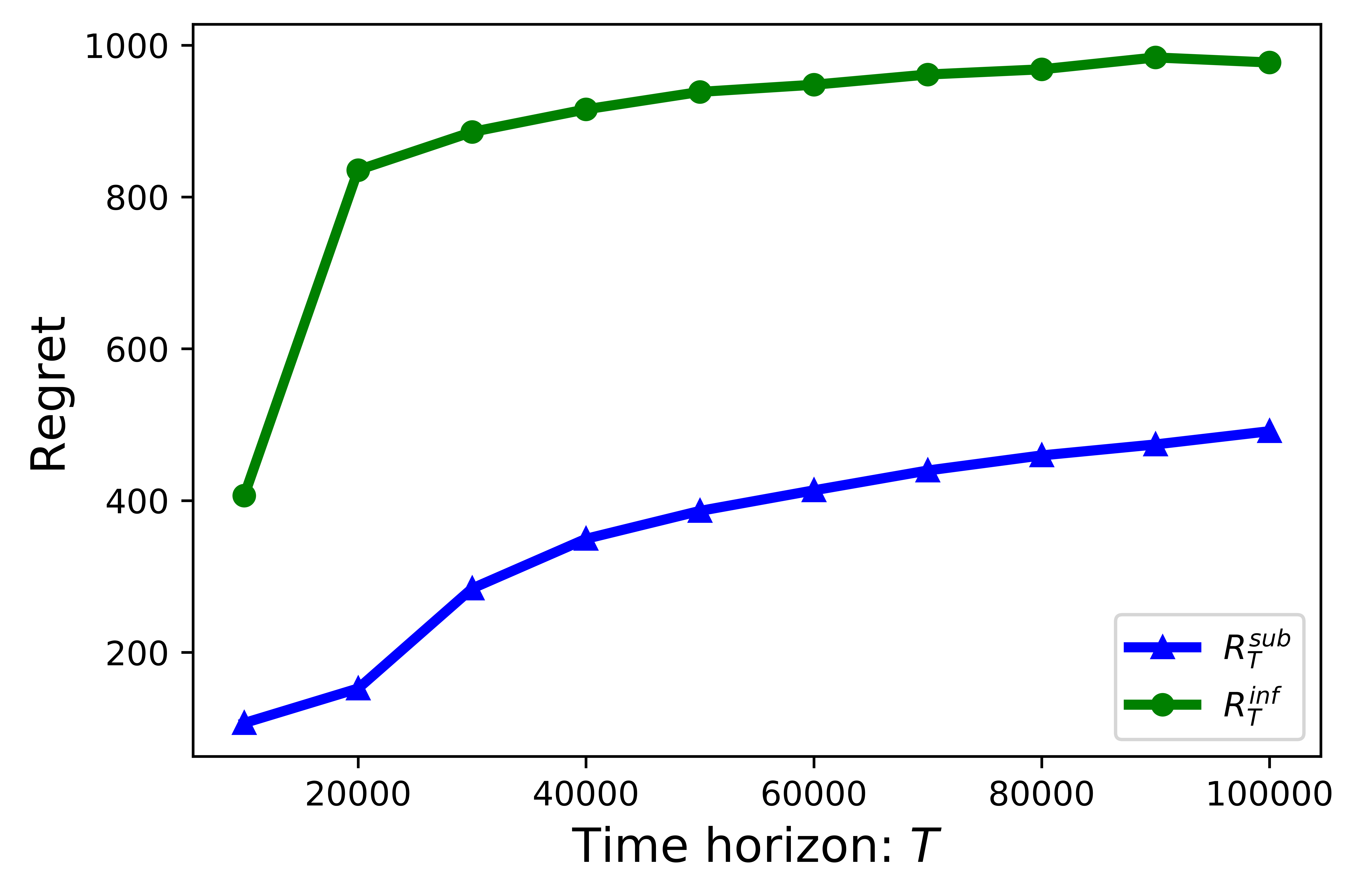}
\caption{Regret of Beta distributed instance}
\label{fig:regret_mv}
\end{figure}

We consider another instance where arms are Beta distributed. The
parameters are chosen in such a way that the mean of the arms
are 0.3, 0.4, 0.2, and 0.5 and the variance of the arms are
0.08, 0.06, 0.15, 0.15. The goal here is to maximize the pulls
of the arm with the minimum mean, subject to a constraint on 
the variance. Specifically, we require that the variance of the
arms should be less than or equal to $\tau := 0.1.$ The instance is
summarized in Table~\ref{tab:mv_remarks}. We use the empirical 
average to estimate the mean of the arms and the standard unbiased
estimator of variance to estimate the variance of the arms.
The concentration bounds we use are based on fact that underlying
arms are bounded between 0 and 1. We run the algorithm 
100 times for values of $T$ in $[10000, 100000].$ The suboptimality regret
$R_{T}^{\text{sub}}$ and the infeasibility regret $R_{T}^{\text{inf}}$
are plotted in Figure~\ref{fig:regret_mv}. Similar to the previous 
case, the algorithm correctly detected feasibility in all of the 100
runs.

\begin{figure}[t]
\centering
\includegraphics[width=0.5\textwidth]{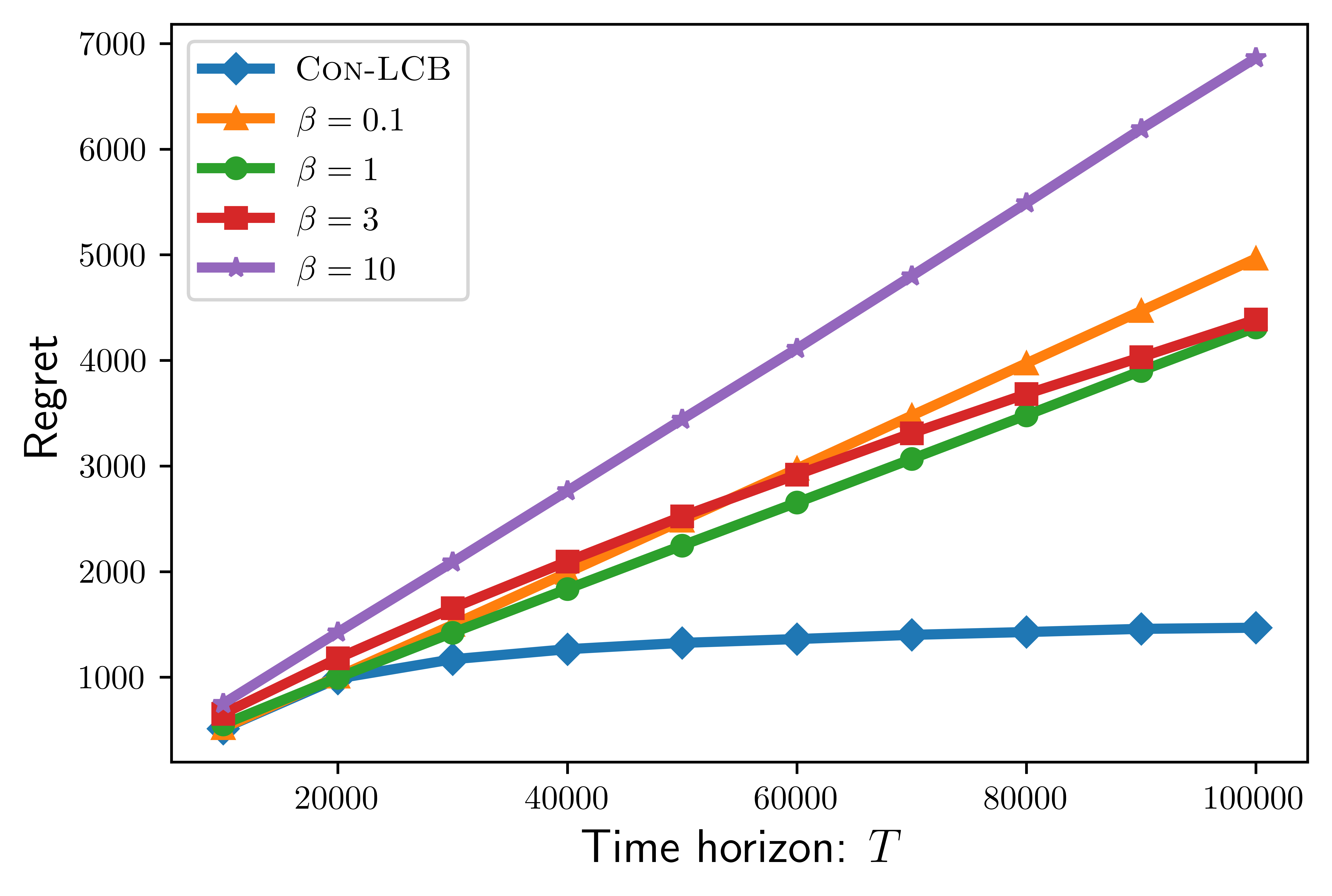}
\caption{Comparison with the Lagrangian relaxation}
\label{fig:lag_lcb}
\end{figure}

\revision{We also compare \textsc{Con-LCB} with an algorithm designed
to optimize a single 'Lagrange relaxed' objective of the form $g_0 + \beta g_1.$
Since there is no systematic way of
setting~$\beta$ such that the original (constrained) optimal arm also
minimizes this metric, this approach is very fragile, as we
demonstrate in Figure~\ref{fig:lag_lcb}. The instance we use is the
Beta distributed instance in Table~\ref{tab:mv_remarks}, 
and we are plotting the sum of the infeasible regret and suboptimal regret. 
When $\beta$ is small ($\beta<10/7$ here), a deceiver arm is optimal 
for the relaxed objective, and when $\beta$ is large ($\beta>5$ here), 
a suboptimal arm is optimal for the relaxed objective; the `correct' range of
$\beta$ being instance dependent (and a priori unknown).
}

\begin{table}[h]
\caption{Beta instance}
\label{tab:mv_remarks}
\begin{tabular}{cccl}
  \textbf{Arm} & \textbf{Mean} & \textbf{Variance} & \textbf{Remarks} \\ 
  1 & 0.3 & 0.08 & Optimal arm \\  
  2 & 0.4 & 0.06 & Feasible, suboptimal \\
  3 & 0.2 & 0.15 & Deceiver \\
  4 & 0.5 & 0.15 & Infeasible, suboptimal 
\end{tabular}
\end{table}

\section{Concluding remarks}
\label{sec:discussion}
In this paper, we have introduced a general formulation of
multi-criterion constrained MABs, which is applicable when there are
multiple attributes of interest associated with each arm. We propose
algorithms that incur logarithmic regret, and also provide information
theoretic lower bounds on the performance of any algorithm. An
interesting departure from the classical MAB formulation is the aspect
of instance feasibility; our algorithms predict, post facto, with a
probability of error that decays as a power law in the horizon,
whether the instance presented was feasible. Interestingly, this
illustrates a fundamental tradeoff between regret minimization and
feasibility detection. Finally, our algorithms `auto-tune' to an
infeasible instance, relaxing the constraints on arm attributes (in
order of increasing importance), until a compliant arm is found.

The proposed framework and our algorithms can be applied in a wide
range of application scenarios (see \cite{Bouneffouf19} for a
survey). Further, this work motivates the study of multi-criterion
variants of other bandit formulations, including contextual bandits,
combinatorial bandits, and correlated bandits.

\section{Declarations}
\label{sec:declarations}
\begin{itemize}
	\item Funding: the authors did not receive support from any organization for the submitted work. 
	\item Conflicts of interest: the authors frequently collaborate with researchers
	from IIT Bombay (@iitb.ac.in), researchers from IIT Madras (@iitm.ac.in), and 
	researchers from Stanford University (@stanford.edu). Jayakrishnan Nair received his
	PhD from Caltech (@caltech.edu) in 2012 and Krishna Jagannathan received his PhD from
	MIT (@mit.edu) in 2010.
	\item Ethics approval: not applicable
	\item Consent to participate: not applicable
	\item Consent for publication: not applicable
	\item Availability of data and material: not applicable
	\item Code availability: the Python code for the numerical experiments can be found in this \href{https://github.com/akagrecha/constrained_regret_minimization}{Github repository}.
	\item Author's contributions: all authors contributed equally to the paper
\end{itemize}

\bibliography{references}   



\newpage
\onecolumn
\appendix

\section{Upper bounds}
\label{app:up_proof}
In this section, we prove Theorems~\ref{thm:con_lcb_feasible1} and \ref{thm:con_lcb_infeasible1}.
The bounds in Subsections~\ref{sec:ub_deceiver}--\ref{sec:prob_mis} imply
the statement of Theorem~\ref{thm:con_lcb_feasible1}, and the
bounds in Subsections~\ref{sec:ub_infeasible}--\ref{sec:prob_mis_infeasible}
imply the statement of Theorem~\ref{thm:con_lcb_infeasible1}.

\subsection{Feasible instance: Upper bounding the expected pulls of
  deceiver arms}
\label{sec:ub_deceiver}
For a deceiver arm $k$, we will define a good event $G_{1,k}$ where the estimator for 
$g_1$ is concentrated enough and derive an upper bound on the number of pulls of the deceiver
arm. 
\begin{equation}
\label{eq:g1_event_def}
G_{1,k} = \left\{ ~\forall n \in [T] ~ |\gest{1}(k) - g_1(k)| <
\sqrt{\frac{\log\left(2T^2\right)}{a_1 n}}  \right\}
\end{equation}

On $G_{1,k},$ we can lower bound the estimator for $g_1$ for the arm $k$ as follows
\begin{equation*}
\gest{1}(k) > g_1(k) - \sqrt{\frac{\log\left(2T^2\right)}{a_1 n}}
\end{equation*}
If the lower bound is greater than $\tau + 
\sqrt{\frac{\log\left(2T^2\right)}{a_1 n}},$ then arm $k$ 
can't be in $\hat{\calk}_t.$ Hence, we can upper bound the number of pulls of an infeasible arm
as follows
\begin{align}
&g_1(k) - \sqrt{\frac{\log\left(2T^2\right)}{a_1 n}} 
\leq \tau + \sqrt{\frac{\log\left(2T^2\right)}{a_1 n}} \nonumber \\
\Rightarrow & n \leq v_k := \ceil*{\frac{4 \log(2T^2)}{a_1 \Delta_{\tau}^2(k)}}
\label{eq:num_g1_bound}
\end{align}

Event $G_{1,k}^c$ is given by
\begin{align*}
G_{1,k}^c = \left\{ ~\exists n \in [T] ~ |\gest{1}(k) - g_1(k)| \geq
\sqrt{\frac{\log\left(2T^2\right)}{a_1 n}}  \right\}
\end{align*}
Using our assumption in Equation~\ref{eq:gen_conc_ineq_m1} and a union bound, we can show
\begin{align*}
\prob{G_{1,k}^c} \leq T \times \frac{1}{T^2} = \frac{1}{T}.
\end{align*}

Now, let us upper bound the expected number of pulls of a deceiver arm $k.$
\begin{align*}
\Exp{N_{k}(T)} &= \Exp{\Exp{N_{k}(T) | G_{1,k}}} + \Exp{\Exp{N_{k}(T) | G_{1,k}^c}} \\
&\leq\ceil*{\frac{4 \log(2T^2)}{a_1 \Delta_{\tau}^2(k)}} \left(1 - \frac{1}{T}\right)
+ T \times \frac{1}{T} \\
&\leq \frac{4 \log(2T^2)}{a_1 \Delta_{\tau}^2(k)} + 2
\end{align*}

\subsection{Feasible instance: Upper bounding the expected pulls of feasible suboptimal arms}
\label{sec:ub_suboptimal}
We will begin by showing that a feasible arm $k$ remains in the set 
$\hat{\calk}_t$ for $t \in [T]$ when estimator of $g_1$ is concentrated enough. We define
an event $G_{1,k}$ for a feasible arm as done in Equation~\eqref{eq:g1_event_def}.
When $G_{1,k}$ holds, the estimator for $g_1$ is upper bounded by
\begin{align*}
\gest{1}(k) \leq & g_1(k) + \sqrt{\frac{\log\left(2T^2\right)}{a_1 n}} \\
\leq & \tau + \sqrt{\frac{\log\left(2T^2\right)}{a_1 n}} \quad (\because~ g_1(k) \leq \tau)
\end{align*} 
Hence, arm $k$ is in $\hat{\calk}_t$ for $t \in [T]$ when $G_{1,k}$ holds.

We are considering the case where an arm $k$ is feasible but
suboptimal. We will define a good event for arm $k$ and bound the
number of pulls on this good event.  Without loss of generality, assume
arm~1 is optimal.

\begin{multline}
\label{eq:g0_event_def}
G_{0,k} = \left\{g_0(1) > \max_{n \in [T]} \gest{0}(1) - 
\sqrt{\frac{\log\left(2T^2\right)}{a_0 n}} \right\} \\
\cap \left\{ \hat{g}_{0, u_k}(k) - 
\sqrt{\frac{\log\left(2T^2\right)}{a_0 u_k}} > g_0(1) \right\} 
\cap G_{1,1} \cap G_{1,k} \\
\text{where } u_k = \ceil*{\frac{4 \log(2T^2)}{a_0 \Delta^2(k)}}
\end{multline}



We will show that if $G_{0,k}$ holds, then $N_{k}(T) \leq u_k$. We will also show that $G_{0,k}^c$ 
holds with a small probability. 

The proof is by contradiction. Suppose $G_{0,k}$ holds and $N_{k}(T) > u_k$, then there exists a 
$t \in [T]$ such that $N_{k}(t-1) = u_k$ and $A_t = k.$ Using the definition of $G_{0,k},$
\begin{align*}
\lcb_{k}(t-1) &= \hat{g}_{0, u_k}(k) - 
\sqrt{\frac{\log\left(2T^2\right)}{a_0 u_k}} \\
& > g_0(1) \qquad (\text{Definition of }G_{0,k}) \\
& > \lcb_{1}(t-1) \qquad (\text{Definition of }G_{0,k})
\end{align*}
Hence, $A_t = \argmin_j \lcb_j (t) \neq k,$ which is a contradiction. Therefore, if $G_{0,k}$
occurs, $N_{k}(T) \leq u_k.$ 

Now, consider the event $G_{0,k}^c.$
\begin{multline*}
G_{0,k}^c = \left\{g_0(1) \leq \max_{n \in [T]} \gest{0}(1) - 
\sqrt{\frac{\log\left(2T^2\right)}{a_0 n}} \right\} \cup \\
\left\{ \hat{g}_{0, u_k}(k) - 
\sqrt{\frac{\log\left(2T^2\right)}{a_0 u_k}} \leq g_0(1) \right\}   \cup G_{1,1}^c \cup G_{1,k}^c.
\end{multline*}

Let us bound the probability of the first term above.
\begin{align*}
&\prob{ g_0(1) \leq \max_{n \in [T]} \gest{0}(1) - 
\sqrt{\frac{\log\left(2T^2\right)}{a_0 n}} } \\
\leq &\prob{\bigcup_{n=1}^{T}  \left\{g_0(1) \leq \gest{0}(1) - 
\sqrt{\frac{\log\left(2T^2\right)}{a_0 n}} \right\}} \\
\leq & T \times \frac{1}{T^2} = \frac{1}{T} \quad 
(\text{Using Equation~\ref{eq:gen_conc_ineq_m1} and union bound})
\end{align*}

Now, let us bound the second event above. By the choice of $u_k$ we have the 
following
\begin{align*}
\Delta(k) - \sqrt{\frac{\log\left(2T^2\right)}{a_0 u_k}}  \geq \sqrt{\frac{\log\left(2T^2\right)}{a_0 u_k}}  
\end{align*}

Now, 
\begin{align*}
&\prob{\hat{g}_{0, u_k}(k) - 
\sqrt{\frac{\log\left(2T^2\right)}{a_0 u_k}} \leq g_0(1)} \\
= &\prob{g_0(k) - \hat{g}_{0,u_k}(k) \geq \Delta(k) - \sqrt{\frac{\log\left(2T^2\right)}{a_0 u_k}}}
\quad (\text{Use } \Delta(k)=g_0(k)-g_0(1)) \\
\leq &\prob{g_0(k) - \hat{g}_{0,u_k}(k) \geq \sqrt{\frac{\log\left(2T^2\right)}{a_0 u_k}}}
\quad (\text{By the choice of } u_k) \\
\leq & \frac{1}{T^2} \quad (\text{Using Equation~}\ref{eq:gen_conc_ineq_m1})
\end{align*}

We can show that $\prob{G_{1,1}^c} \leq \nicefrac{1}{T}$ and $\prob{G_{1,k}^c} \leq \nicefrac{1}{T}$
like in the previous subsection. Hence,
\begin{align*}
	\prob{G_{0,k}^c} \leq \frac{3}{T} + \frac{1}{T^2}.
\end{align*}

Hence, we can upper bound the number of pulls of feasible but suboptimal arms as follows
\begin{align*}
\Exp{N_{k}(T)} = &\Exp{\Exp{N_{k}(T) | G_{0,k}}} + \Exp{\Exp{N_{k}(T) | G_{0,k}^c}} \\
\leq &\ceil*{\frac{4 \log(2T^2)}{a_0 \Delta^2(k)}} \left(1 - \frac{3}{T} - \frac{1}{T^2} \right) +
T \times \left(\frac{3}{T} + \frac{1}{T^2} \right) \\
\leq &\frac{4 \log(2T^2)}{a_0 \Delta^2(k)} + 5.
\end{align*}

\subsection{Feasible instance: Upper bounding the expected pulls of infeasible suboptimal arms}
\label{sec:ub_infeasible_suboptimal}
Consider arm $k$ which is both suboptimal and infeasible. Define an event $G_{0,k}$ as done in
Equation~\eqref{eq:g0_event_def}. Recall that the upper bound on the pulls of the deceiver arms
on event $G_{1,k}$ is denoted by $v_k$ and the upper bound on the pulls of the feasible but
suboptimal arms on event $G_{0,k}$ is denoted by $u_k.$ 

On the event $G_{0,k},$ if $u_k \geq v_k,$ then due to concentration of estimator of $g_1$, 
this arm can't be played more than $v_k$ times. If $u_k < v_k,$ then due to suboptimality, 
this arm can't be played more than $u_k$ times. We can show that the probability of 
$G_{0,k}^c$ is less than or equal to $\nicefrac{3}{T} + \nicefrac{1}{T^2}$ as we did before.
Hence, we can upper bound the pulls of infeasible and suboptimal arms as follows
\begin{align*}
\Exp{N_{k}(T)} = &\Exp{\Exp{N_{k}(T) | G_{0,k}}} + \Exp{\Exp{N_{k}(T) | G_{0,k}^c}} \\
& \leq \min(u_k, v_k) \left(1 - \frac{3}{T} - \frac{1}{T^2}\right) + 
T \times \left( \frac{3}{T} + \frac{1}{T^2} \right) \\
& \leq \min(u_k, v_k) + 4
\end{align*}

\subsection{Feasible instance: Upper bounding the probability of misidentification}
\label{sec:prob_mis}
Feasibility is correctly detected if for at least one of the feasible arms, event $G_{1,k}$ as defined
in Equation~\eqref{eq:g1_event_def} holds. We had seen in Subsection~\ref{sec:ub_suboptimal}
that if estimator of $g_1$ is concentrated enough, a feasible arm always remains in the 
set $\hat{\calk}_t$ of plausibly feasible arms.

Without loss of generality, assume that arm 1 is optimal. Then, we can lower bound the probability of 
correctly setting the flag as follows
\begin{align*}
\prob{\texttt{feasibility\_flag = true}} &\geq \prob{G_{1,1}} \\
&\geq 1 - \frac{1}{T}. 
\end{align*}
This upper bounds the probability of incorrectly setting the flag
\begin{equation*}
\prob{\texttt{feasibility\_flag = false}} \leq \frac{1}{T}.
\end{equation*}

\subsection{Infeasible instance: Upper bounding the expected pulls of non-optimal arms}
\label{sec:ub_infeasible}
In this section, we discuss the case when the given instance is infeasible. As defined before,
the optimal choice for the algorithm is to play the arm with minimum $g_1$. We will upper bound 
the number of pulls of arms that have $g_1$ greater than the minimum. 

For all the arms, we define good events $G_{1,k}$ as in Equation~\eqref{eq:g1_event_def}
where the estimator for $g_1$ is concentrated enough. Let $\mathcal{E}_r = \cap_{k=1}^{K} G_{1,k}.$ 
When event $\mathcal{E}_r$ occurs, the set $\hat{\calk}_t$ becomes empty after at 
most $\sum_{k=1}^{K} v_k$ pulls where $v_k$ is defined in Equation~\eqref{eq:num_g1_bound}. 
The analysis is similar to given in Subsection~\ref{sec:ub_deceiver}. 

Once the set $\hat{\calk}_t$ becomes empty, the algorithm starts pulling arms with minimum 
lower confidence bound on estimator of $g_1$. We will upper bound the number of pulls for the 
non-optimal arms. Without loss of generality, assume that arm 1 has the lowest $g_1$. 
As we are dealing with an infeasible instance, $g_1(1)>\tau.$ For a non-optimal arm $k,$ 
we define the following good event
\begin{align*}
G_{r,k} = \left\{ \hat{g}_{1, w_k} - \sqrt{\frac{\log\left(2T^2\right)}{a_1 w_k}} 
> g_1(1) \right\} \cap \mathcal{E}_r \\
\text{where } w_k = \ceil*{\frac{4 \log(2T^2)}{a_1 \Delta_{\text{con}}^2(k)}}.
\end{align*}

One can check that $w_k$ is greater than $v_k$ because the gap 
$\Delta_{\text{con}}(k) = g_1(k) - g_1(1)$ is smaller than  $\Delta_\tau(k) = g_1(k) - \tau.$ 
It is easy to argue using LCB based arguments that if $G_{r,k}$ occurs, then arm $k$ can't
be pulled more than $w_k$ times. This is similar to the proof given in 
Subsection~\ref{sec:ub_suboptimal}.  

Let us upper bound the probability of $G_{r,k}^c.$ Event $G_{r,k}^c$ is given by
\begin{align*}
G_{r,k}^c = \left\{ \hat{g}_{1, w_k} - \sqrt{\frac{\log\left(2T^2\right)}{a_1 w_k}} 
\leq g_1(1) \right\}  \cup \mathcal{E}_r^c.
\end{align*}
Using analysis in Subsection~\ref{sec:ub_deceiver}, we can show 
$\prob{\mathcal{E}_r^c} \leq \nicefrac{K}{T}.$ 
Let us bound the probability of the first term above. By the choice of $w_k,$ we have
\begin{align*}
\Delta_{\text{con}}(k) - \sqrt{\frac{\log\left(2T^2\right)}{a_1 w_k}} \geq
\sqrt{\frac{\log\left(2T^2\right)}{a_1 w_k}}
\end{align*}

Now,
\begin{align*}
&\prob{\hat{g}_{1, w_k} - \sqrt{\frac{\log\left(2T^2\right)}{a_1 w_k}} \leq g_1(1)} \\
= &\prob{g_1(k) - \hat{g}_{1, w_k} \geq \Delta_{\text{con}}(k) - \sqrt{\frac{\log\left(2T^2\right)}{a_1 w_k}}}
\quad (\text{Use } \Delta_{\text{con}}(k) = g_1(k) - g_1(1)) \\
\leq &\prob{g_1(k) - \hat{g}_{1, w_k} \geq \sqrt{\frac{\log\left(2T^2\right)}{a_1 w_k}}} 
\quad (\text{By the choice of } w_k)\\
\leq &\frac{1}{T^2} \quad (\text{Using Equation~}\ref{eq:gen_conc_ineq_m1}) 	
\end{align*}

Hence, we can upper bound $\prob{G_{r,k}^c}$ using a union bound
\begin{align*}
\prob{G_{r,k}^c} \leq \frac{K}{T} + \frac{1}{T^2} \leq \frac{K+1}{T}.
\end{align*}

When the instance is infeasible, the expected number of pulls of non-optimal arms are upper bounded by 
\begin{align*}
\Exp{N_{k}(T)} &= \Exp{\Exp{N_{k}(T) | G_{r,k}}} +  \Exp{\Exp{N_{k}(T) | G_{r,k}^c}} \\
& \leq \ceil*{\frac{4 \log(2T^2)}{a_1 \Delta_{\text{con}}^2(k)}} \left(1 - \frac{K+1}{T} \right)	
+ T \times \frac{K+1}{T} \\
& \leq \frac{4 \log(2T^2)}{a_1 \Delta_{\text{con}}^2(k)} + K+2
\end{align*} 

\subsection{Infeasible instance: Upper bounding the probability of misidentification}
\label{sec:prob_mis_infeasible}
In this subsection, we will upper bound the probability of incorrectly setting the \texttt{feasibility\_flag}.

Firstly, we define $T^*$ to be the minimum value of $T$ for which the following holds:
\begin{align*}
T > \sum_{k=1}^{K} v_{k} \geq \sum_{k=1}^{K} \ceil*{\frac{4 \log(2T^2)}{a_1 \Delta_{\tau}^2(k)}}.
\end{align*}
$T^*$ exists because $C\log(T) = o(T)$ for a fixed $C \in (0, \infty).$

For all the arms, we define good events $G_{1,k}$ as in Equation~\eqref{eq:g1_event_def}
where the estimator for $g_1$ is concentrated enough. Let $\mathcal{E}_r = \cap_{k=1}^{K} G_{1,k}.$ 
On event $\mathcal{E}_r,$ for $t>T^*,$ set of plausible feasible arms $\hat{\calk}_t$ 
will remain empty.

We showed in the previous subsection that $\prob{\mathcal{E}_r} \geq 1 - \frac{K}{T}.$
Hence, we can bound the probability of incorrectly setting the \texttt{feasibility\_flag} as
\begin{align*}
\prob{\texttt{feasibility\_flag = true}} \leq \frac{K}{T} \text{ for } T>T^*.  	
\end{align*}


\section{Lower bounds}
\label{app:lower_bound}

In this section, we will prove Theorems~\ref{thm:lower_bound_gaussian} and 
\ref{thm:prob_incorrect_flag}. To prove Theorem~\ref{thm:lower_bound_gaussian}, 
we will first prove a result which lower bounds the pulls of non-optimal arms
when the arm distributions belong to a general distribution class and the 
attributes are not necessarily means. Theorem~\ref{thm:lower_bound_gaussian}
is proved in Subsection~\ref{sec:gaussian_pulls},  Theorem~\ref{thm:prob_incorrect_flag} 
is proved in Subsection~\ref{sec:disambiguate}, and subsequent subsections provide the proof
for the intermediate results.  

The proof technique to derive lower bounds for the constrained bandit setting is very similar
to the technique used for standard stochastic bandit setting. We begin by stating some
important results that will be used later in the proofs.

We first state the divergence decomposition lemma for the constrained bandit setting.
The proof of the lemma is similar to the proof of divergence decomposition for the 
standard setting and we leave it to the reader to verify the result (see Lemma~15.1,
\cite{lattimore2018}).    

\begin{lemma}
Consider two instances $(\nu, \tau)$ and $(\nu', \tau),$ where $\nu$ and $\nu'$ belong
to a space of distributions $\mathcal{C}^K.$ Fix some policy $\pi$ and let 
$\mathbb{P}_{\nu} = \mathbb{P}_{(\nu, \tau), \pi}$ and $\mathbb{P}_{\nu'} = \mathbb{P}_{(\nu',\tau) \pi}$ 
be the probability measures on the constrained bandit model induced by the $T$-round interconnection between
$\pi$ and $(\nu, \tau)$ (respectively, $\pi$ and $(\nu', \tau)$). Then
\begin{align}
	\KL(\mathbb{P}_{\nu}, \mathbb{P}_{\nu'}) = \sum_{k=1}^{K} \Expnu{\nu}{N_k(T)} \KL(P_i, P_i')
\end{align}
\end{lemma} 

We also state the high probability Pinsker inequality (see Theorem~14.2, \cite{lattimore2018}). 
\begin{lemma}
\label{lem:high_prob_pinsker}
Let $P$ and $Q$	be probability measures on the same measurable space $(\Omega, \mathcal{F})$ and let 
$A \in \mathcal{F}$ be an arbitrary event. Then,
\begin{align*}
P(A) + Q(A^c) \geq \frac{1}{2}\exp(-\KL(P,Q)),	
\end{align*}
where $A^c = \Omega / A$ is the complement of A.
\end{lemma}

\subsection{Lower bounding the pulls of non-optimal arms for the Gaussian case}
\label{sec:gaussian_pulls}
We first state an instance-dependent lower bound for the expected
pulls of non-optimal arms under consistent algorithms for any 
class of distributions $\mathcal{C}$ and a
threshold~$\tau \in \R.$ For a feasible instance~$(\nu^f,\tau),$ where
$\nu^f \in \mathcal{C}^K,$ define, for each non-optimal arm~$k,$
\begin{equation*}
\eta^{f}(\nu^f(k), g_0^*, \tau, \mathcal{C}) = 
\inf_{\nu'(k) \in \mathcal{C}} \{ \KL(\nu^f(k), \nu'(k)) : g_0(\nu'(k))<g_0^*, ~g_1(\nu'(k))\leq \tau \}.
\end{equation*}
Similarly, for an infeasible instance $(\nu^i,\tau),$ where
$\nu' \in \mathcal{C}^K,$ define, for each non-optimal arm~$k,$
\begin{equation*}
\eta^{i}(\nu^i(k), g_1^*, \mathcal{C}) = \inf_{\nu'(k) \in \mathcal{C}}\{\KL(\nu^i(k), \nu'(k)) : g_1(\nu'(k)) < g_1^*\}.  	
\end{equation*}
\begin{theorem}
\label{thm:lower_bound_pulls}
Let $\pi$ be a consistent policy over the class of distributions
$\mathcal{C}$ given the threshold $\tau \in \mathbb{R}.$ Then for
a feasible instance $(\nu^f,\tau),$ where $\nu^f \in \mathcal{C}^K,$
for any non-optimal arm $k,$ 
\begin{align*}
\liminf_{T \to \infty} \frac{\Expnu{\nu}{N_{k}(T)}}{\log(T)} \geq \frac{1}{\eta^{f}(\nu^f(k), g_0^*, \tau, \mathcal{C})},
\end{align*}   	
For an infeasible instance $(\nu^i,\tau),$ where
$\nu' \in \mathcal{C}^K,$ for any non-optimal arm $k,$
\begin{align*}
\liminf_{T \to \infty} \frac{\Expnu{\nu}{N_{k}(T)}}{\log(T)} \geq \frac{1}{\eta^{i}(\nu^i(k), g_1^*, \mathcal{C})}.
\end{align*}
\end{theorem}
The bounds in Subsections~\ref{sec:proof_feasible} and
\ref{sec:proof_infeasible} imply Theorem~\ref{thm:lower_bound_pulls}.
The main message here is that the 'hardness' of a non-optimal arm~$k$, 
which dictates the minimum number of pulls required on average to 
distinguish it from the optimal arm, is characterized by the reciprocal
of the smallest perturbation in terms of KL divergence $\KL(\nu(k),\cdot)$ 
needed to `make' the arm optimal.

Now, we will use Theorem~\ref{thm:lower_bound_pulls} to prove 
Theorem~\ref{thm:lower_bound_gaussian}. 

Consider two d-dimensional Gaussian distribution $\nu_1$ and $\nu_2$ with means 
$\mu_1$ and $\mu_2,$ and covariances $\Sigma_1$ and $\Sigma_2.$ The KL divergence 
between these distributions is given by
\begin{equation}
\label{eq:kl_g_gen}
\KL(\nu_1, \nu_2) = \log\left(\frac{|\Sigma_2|^{0.5}}{|\Sigma_1|^{0.5}}\right)
+ \frac{\text{tr}(\Sigma_2^{-1} \Sigma_1) -\text{tr} (\Sigma_1^{-1}\Sigma_1) + (\mu_1 - \mu_2)^T\Sigma_2^{-1}(\mu_1 - \mu_2)}{2}.
\end{equation}
This is easy to derive but one can check Section~9 of these \href{https://stanford.edu/~jduchi/projects/general_notes.pdf}{notes}.

Let $\mathcal{G}$ be the class of 2-dimensional gaussian distributions with the covariance
matrix $\Sigma = \text{diagonal}(\nicefrac{1}{2 a_0}, \nicefrac{1}{2 a_1}).$ Let attribute 
$g_0$ be the mean of the first dimension, and $g_1$ be the mean of the second dimension. 
Let $\nu_1, \nu_2 \in \mathcal{G},$ then using Equation~\ref{eq:kl_g_gen}, the KL divergence
between $\nu_1$ and $\nu_2$ is given by
\begin{equation}
\label{eq:kl_g}
 \KL(\nu_1, \nu_2) = a_0 (g_0(\nu_1) - g_0(\nu_2))^2 + a_1 (g_1(\nu_1) - g_1(\nu_2))^2. 
\end{equation} 

With $\mathcal{C} = \mathcal{G}$ and attributes as defined in 
Theorem~\ref{thm:lower_bound_gaussian}, we can use Equation~\ref{eq:kl_g} 
to calculate $\eta^{f}(\nu^f(k), g_0^*, \tau, \mathcal{C})$
and $\eta^{i}(\nu^i(k), g_1^*, \mathcal{C}).$ In particular, for a
feasible but suboptimal arm $k,$ we have
\begin{align*}
\eta^{f}(\nu^f(k), g_0^*, \tau, \mathcal{C}) = a_0 \Delta^2(k),
\end{align*}
for a deceiver arm $k,$ we have
\begin{align*}
\eta^{f}(\nu^f(k), g_0^*, \tau, \mathcal{C}) = a_1 [\Delta_\tau(k)]^2,
\end{align*}
and for an infeasible arm $k$ which is not a deceiver, we have
\begin{align*}
\eta^{f}(\nu^f(k), g_0^*, \tau, \mathcal{C}) = a_0 \Delta^2(k) + a_1 [\Delta_\tau(k)]^2
\leq 2 \max(a_0 \Delta^2(k), a_1 [\Delta_\tau(k)]^2).
\end{align*}  
Finally, for a non-optimal arm $k$ for an infeasible instance, we have
\begin{align*}
\eta^{i}(\nu^i(k), g_1^*, \mathcal{C}) = a_1 [\Delta_\text{con}(k)]^2. 	
\end{align*} 
Using the results above and combining them with Theorem~\ref{thm:lower_bound_pulls},
we get Theorem~\ref{thm:lower_bound_gaussian}.


\subsection{Disambiguating between feasible and infeasible instances}
\label{sec:disambiguate}
Consider a feasible instance $(\nu, \tau)$ where Arm~1 is the only feasible arm and therefore,
the only optimal arm. Let $g_1^{\dagger} = \min_{k \in \{2,\cdots,K\}}g_1(k)$ be the minimum value
of $g_1$ for the set of infeasible arms. For Arm~1 we define
\begin{align*}
\eta(\nu(1), g_1^{\dagger}, \mathcal{C}) = 
\inf_{\nu'(1) \in \mathcal{C}}\{\KL(\nu'(1), \nu(1)): g_1(\nu'(1))>g_1^{\dagger}\}.     	
\end{align*}    
Consider another instance $(\nu', \tau)$ where $\nu'(j) = \nu(j)$ for $j \neq 1$ and 
$\nu'(1) \in \mathcal{C}$ such that $\KL(\nu(1), \nu'(1)) \leq d_1 + \varepsilon$ and $g_1(\nu'(1)) > g_1^{\dagger},$
where $d_1 = \eta(\nu(1), g_1^{\dagger}, \mathcal{C}).$ Using divergence decomposition lemma, 
we have $\KL(\mathbb{P}_{\nu'}, \mathbb{P}_{\nu}) \leq \Expnu{\nu' }{N_1(T)}(d_1+\varepsilon)$ and by using 
Lemma~\ref{lem:high_prob_pinsker} we have
\begin{align*}
\probnu{\nu}{A} + \probnu{\nu'}{A^c} \geq \frac{1}{2}\exp(-\KL(\mathbb{P}_{\nu'}, \mathbb{P}_{\nu}))	
\geq \frac{1}{2} \exp(-\Expnu{\nu' }{N_1(T)}(d_1+\varepsilon)).
\end{align*}
Let event $A =$ \{\texttt{feasibility\_flag} = false\}. Taking logarithm on both sides and
rearranging gives
\begin{align*}
-\frac{\log(\probnu{\nu}{A} + \probnu{\nu'}{A^c})}{T} \leq 
\frac{\log(2) + (d_1 + \varepsilon)\Expnu{\nu' }{N_1(T)}}{T} \\
\end{align*}
RHS goes to zero as $T$ goes to infinity. This follows from the definition of consistency and the fact that 
for instance $(\nu', \tau),$ arm~1 is suboptimal. Hence, we have
\begin{align*}
\limsup_{T \to \infty} -\frac{\log(\probnu{\nu}{A} + \probnu{\nu'}{A^c})}{T} \leq 0
\end{align*}
This shows that at least for one of the instances, the probability of incorrect detection decays slower 
than exponential in $T.$ 

\subsection{Feasible instances}
\label{sec:proof_feasible}
Consider a class of distributions $\mathcal{C}$ and a 
threshold~$\tau \in \R.$ For a feasible instance~$(\nu^f,\tau),$ where
$\nu^f \in \mathcal{C}^K,$ define, for each non-optimal arm~$k,$
\begin{equation*}
\eta^{f}(\nu^f(k), g_0^*, \tau, \mathcal{C}) = 
\inf_{\nu'(k) \in \mathcal{C}} \{ \KL(\nu^f(k), \nu'(k)) : g_0(\nu'(k))<g_0^*, ~g_1(\nu'(k))\leq \tau \}.
\end{equation*}

We will show that
\begin{equation*}
\liminf_{T \to \infty} \frac{\Expnu{\nu}{N_k(T)}}{\log(T)} \geq \frac{1}{\eta^{f}(\nu^f(k), g_0^*, \tau, \mathcal{C})}.
\end{equation*}
\begin{proof}
Let $d_k = \eta^{f}(\nu^f(k), g_0^*, \tau, \mathcal{C})$ and fix any $\varepsilon > 0.$ Let $(\nu', \tau)$ 
be a bandit instance with $\nu' \in \mathcal{C}^K,$ and $\nu'(j) = \nu^f(j)$ for $j \neq k$ 
be such that $\KL(\nu^f(k), \nu'(k)) \leq d_k + \varepsilon,$ $g_0(\nu'(k))<g_0^*,$ 
and $g_1(\nu'(k)) \leq \tau.$ A distribution like $\nu'(k)$ exists because of the definition 
of $d_k.$ Note that arm $k$ is the unique optimal arm for bandit instance $\nu'.$ 
Using divergence decomposition lemma, we have $\KL(\mathbb{P}_{\nu^f}, \mathbb{P}_{\nu'}) 
\leq \Expnu{\nu}{N_k(T)}(d_k+\varepsilon)$ and by using Lemma~\ref{lem:high_prob_pinsker} we have
\begin{align*}
\probnu{\nu}{A} + \probnu{\nu'}{A^c} \geq \frac{1}{2}\exp(-\KL(\mathbb{P}_{\nu^f}, \mathbb{P}_{\nu'}))	
\geq \frac{1}{2} \exp(-\Expnu{\nu^f}{N_k(T)}(d_k+\varepsilon)).
\end{align*}
Let event $A = \{N_k(T) > \nicefrac{T}{2}\}.$ 
\begin{align*}
\Expnu{\nu^f}{N_k(T)} + \sum_{j \neq k} \Expnu{\nu'}{N_{j}(T)} &\geq 
\frac{T}{2} (\probnu{\nu^f}{A} + \probnu{\nu'}{A^c}), \\
&\geq \frac{T}{4} \exp(-\Expnu{\nu^f}{N_k(T)}(d_k+\varepsilon)).
\end{align*}
Rearranging and taking the limit inferior we get
\begin{align*}
\liminf_{T \to \infty} \frac{\Expnu{\nu^f}{N_k(T)}}{\log T} &\geq
\frac{1}{d_k + \varepsilon} \liminf_{T \to \infty} 
\frac{\log \left(\frac{T}{4(\Expnu{\nu^f}{N_k(T)} + \sum_{j \neq k}\Expnu{\nu'}{N_{j}(T)})} \right)}{\log T} \\
& = \frac{1}{d_k+\varepsilon} \left(1 - \limsup_{T\to\infty} \frac{\log(\Expnu{\nu^f}{N_k(T)} + \sum_{j \neq k}\Expnu{\nu'}{N_{j}(T)})}{\log T} \right) \\
& = \frac{1}{d_k + \varepsilon}.
\end{align*} 
The last equality follows from the definition of consistency. As an arm which is suboptimal or
infeasible or both is played only $o(T^a)$ times in expectation for all $a>0$, there exists a constant $C_a$
for large enough $T$ such that 
$\Expnu{\nu^f}{N_k(T)} + \sum_{j \neq k}\Expnu{\nu'}{N_{j}(T)}) \leq C_a T^a.$
This gives
\begin{align*}
\limsup_{T\to\infty} \frac{\log(\Expnu{\nu^f}{N_k(T)} + \sum_{j \neq k}\Expnu{\nu'}{N_{j}(T)})}{\log T} 
\leq \limsup_{T \to\infty} \frac{C_a + a \log(T)}{\log(T)} = a.
\end{align*}
As this holds for all $a>0$ and $\varepsilon$ was an arbitrary constant greater than zero,
we have the result.
\end{proof}

\subsection{Infeasible instance}
\label{sec:proof_infeasible}
For an infeasible instance $(\nu^i,\tau),$ where
$\nu' \in \mathcal{C}^K,$ define, for each non-optimal arm~$k,$
\begin{equation*}
\eta^{i}(\nu^i_k, g_1^*, \mathcal{C}) = \inf_{\nu'(k) \in \mathcal{C}}\{\KL(\nu^i_k, \nu'(k)) : 
g_1(\nu'(k)) < g_1^*\}.  	
\end{equation*}
We will show that for arm $k$
\begin{align*}
\liminf_{T \to \infty} \frac{\Expnu{\nu}{N_k(T)}}{\log(T)} \geq \frac{1}{\eta^{i}(\nu^i_k, g_1^*, \mathcal{C})}	
\end{align*}
\begin{proof}
Let $d_k = \eta^{i}(\nu^i_k, g_1^*, \mathcal{C})$ and fix any $\varepsilon>0$. 
Let $(\nu', \tau)$ be a bandit instance with $\nu'(j) = \nu^i(j)$ for $j \neq k$ 
and $\nu'(k) \in \mathcal{C}$ such that $\KL(\nu^i(k), \nu'(k)) \leq d_k + \varepsilon$ 
and $g_1(\nu'(k)) \leq g_1^*.$ A distribution like $\nu'(k)$ exists because of the 
definition of $d_k.$ Observe that the instance $(\nu', \tau)$ could be a feasible instance. 
Nonetheless, arm $k$ is the unique optimal arm irrespective of the 
feasibility of instance $(\nu', \tau).$ Using divergence decomposition lemma, 
we have $\KL(\mathbb{P}_{\nu^i}, \mathbb{P}_{\nu'}) \leq \Expnu{\nu^i}{N_k(T)}(d_k+\varepsilon)$ 
and by using Lemma~\ref{lem:high_prob_pinsker} we have
\begin{align*}
\probnu{\nu^i}{A} + \probnu{\nu'}{A^c} \geq \frac{1}{2}\exp(-\KL(\mathbb{P}_{\nu^i}, \mathbb{P}_{\nu'}))	
\geq \frac{1}{2} \exp(-\Expnu{\nu^i}{N_k(T)}(d_k+\varepsilon)).
\end{align*}
Let event $A = \{N_k(T) > \nicefrac{T}{2}\}.$  
\begin{align*}
\Expnu{\nu^i}{N_k(T)} + \sum_{j \neq k} \Expnu{\nu'}{N_{j}(T)} &\geq 
\frac{T}{2} (\probnu{\nu^i}{A} + \probnu{\nu'}{A^c}), \\
&\geq \frac{T}{4} \exp(-\Expnu{\nu^i}{N_k(T)}(d_k+\varepsilon)).
\end{align*}
Rearranging and taking the limit inferior we get
\begin{align*}
\liminf_{T \to \infty} \frac{\Expnu{\nu^i}{N_k(T)}}{\log T} &\geq
\frac{1}{d_k + \varepsilon} \liminf_{T \to \infty} 
\frac{\log \left(\frac{T}{4(\Expnu{\nu^i}{N_k(T)} + \sum_{j \neq k}\Expnu{\nu'}{N_{j}(T)})} \right)}{\log T} \\
& = \frac{1}{d_k+\varepsilon} \left(1 - \limsup_{T\to\infty} \frac{\log(\Expnu{\nu^i}{N_k(T)} + \sum_{j \neq k}\Expnu{\nu'}{N_{j}(T)})}{\log T} \right) \\
& = \frac{1}{d_k + \varepsilon}.
\end{align*} 
The last equality follows from the definition of consistency. As $\nu^i$ is an infeasible instance, 
arm $k$ is suboptimal and is played only $o(T^a)$ times in expectation for all $a>0.$ 
For instance $\nu',$ arm $k$ is the unique optimal arm. Therefore, all the other arms are 
played only $o(T^a)$ times in expectation for all $a>0.$ Hence, there exists a constant $C_a$ 
for large enough $T$ such that $\Expnu{\nu^i}{N_k(T)} + \sum_{j \neq k}\Expnu{\nu'}{N_{j}(T)}) \leq C_a T^a.$ This gives
\begin{align*}
\limsup_{T\to\infty} \frac{\log(\Expnu{\nu^i}{N_k(T)} + \sum_{j \neq k}\Expnu{\nu'}{N_{j}(T)})}{\log T} 
\leq \limsup_{T \to\infty} \frac{C_a + a \log(T)}{\log(T)} = a.
\end{align*}
As this holds for all $a>0$ and $\varepsilon$ was an arbitrary constant greater than zero,
we have the result.
\end{proof}



\end{document}